\documentclass[10pt,twocolumn,letterpaper]{article}

\usepackage{cvpr}
\usepackage{times}
\usepackage{epsfig}
\usepackage{graphicx}
\usepackage{amsmath}
\usepackage{amssymb}

\usepackage{algorithm}
\usepackage{algpseudocode}
\usepackage{mathtools}
\usepackage{subfigure}
\usepackage{colortbl}
\usepackage{enumitem}
\usepackage{xfrac}
\usepackage{appendix}

\newcommand{\namedirect}{Ours (direct pred.)}
\newcommand{\nameavg}{Ours (uniform weight)}
\newcommand{\namekpn}{Ours (kernel pred.)}
\newcommand{\namemodel}{Our Model}
\newcommand{\namepwc}{PWC-Net~\cite{Sun2018PWCNet}}
\newcommand{\nameepic}{EpicFlow~\cite{EpicFlow}}
\newcommand{\namesepconv}{SepConv~\cite{NiklausICCV2017}}
\newcommand{\nameslomo}{Super SloMo~\cite{SuperSlomo2018}}
\newcommand{\namebaseline}{Naive Baseline}

\usepackage[pagebackref=true,breaklinks=true,letterpaper=true,bookmarks=false]{hyperref}

\cvprfinalcopy 

\def\cvprPaperID{1607} 

\ifcvprfinal\fi
\begin{document}

\title{Learning to Synthesize Motion Blur}

\author{
\begin{tabular}{c@{\hspace{0.3in}}c}
Tim Brooks
&
Jonathan T. Barron
\end{tabular}
\\
Google Research
\vspace{0.1in}
}

\maketitle

\begin{abstract}
We present a technique for synthesizing a motion blurred image from a pair of unblurred images captured in succession. To build this system we motivate and design a differentiable ``line prediction'' layer to be used as part of a neural network architecture, with which we can learn a system to regress from image pairs to motion blurred images that span the capture time of the input image pair. Training this model requires an abundance of data, and so we design and execute a strategy for using frame interpolation techniques to generate a large-scale synthetic dataset of motion blurred images and their respective inputs. We additionally capture a high quality test set of real motion blurred images, synthesized from slow motion videos, with which we evaluate our model against several baseline techniques that can be used to synthesize motion blur. Our model produces higher accuracy output than our baselines, and is significantly faster than baselines with competitive accuracy.
\end{abstract}

\maketitle

\section{Introduction}

\begin{figure}[b!]
  \centering
  \subfigure[A pair of input images.]{
  \label{fig:teaser-a}
  \begin{tabular}{@{}c@{}}
  \includegraphics[width=0.98\linewidth]{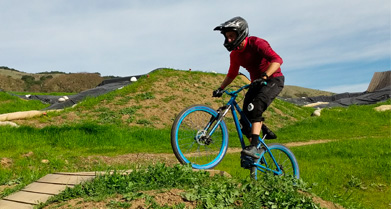} \\
  \includegraphics[width=0.98\linewidth]{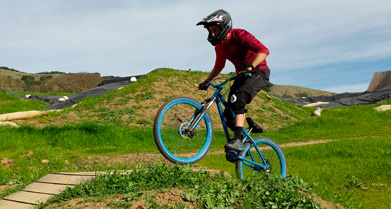}
  \end{tabular}
  }
  \subfigure[Our model's output.]{
  \includegraphics[width=0.98\linewidth]{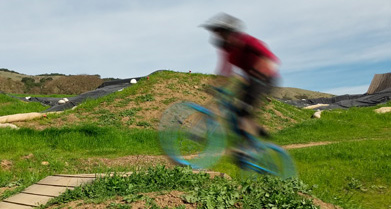}
  \label{fig:teaser-b}
  }
  \vspace{0.05in}
  \caption{In \subref{fig:teaser-a} we present two images of a subject moving across the image plane. Our system uses these images to synthesize the motion blurred image in \subref{fig:teaser-b}, which conveys a sense of motion and separates the subject from the background.
  \label{fig:teaser}
  }
\end{figure}

Though images are often thought of as capturing a single moment in time, all images in fact capture a duration of time: an image begins when a camera starts collecting light, and ends when that camera stops collecting light.
If the camera or the scene move while light is being collected, the resulting image will exhibit motion blur. That blur may indicate the speed of a subject or may serve to separate a subject from the background, depending on the relative motion of the camera and the subject (see Figure~\ref{fig:teaser-b}).

Motion blur is a valuable cue for image understanding.
Given a single image containing motion blur, one can estimate the relative direction and magnitude of scene motion that resulted in the observed blur \cite{Chakrabarti2010AnalyzingSB,MotionFromBlur}. This motion estimate may be semantically meaningful~\cite{WalkerGH15}, or may be used by a deblurring algorithm to synthesize a sharp image \cite{Bascle1996,Fergus06,Levin2006,nayar2004motion}.
Recent work has relied on deep learning for removing motion blur and inferring the underlying motion of the scene \cite{Chakrabarti16,gong2017blur2mf,Sun2015LearningAC}. Deep learning techniques tend to need an abundance of training data to work well, and so to train these techniques one must generate large amounts of synthetic training data by synthetically blurring sharp images. These techniques also tend to use synthetic data (usually sharp images convolved by real or synthetic ``camera shake'' kernels) for quantitative evaluation, using real motion-blurred images only to produce qualitative visualizations.
Naturally, the ability of these learned models to generalize to real images depends on the realism of their synthetic training data.
In this paper, we treat the \emph{inverse} of this well-studied blur removal task as a first class problem. 
We present a fast and effective way to synthesize the training data necessary to train a motion deblurring algorithm, and we quantitatively demonstrate that our technique generalizes from our synthetic training data to real motion-blurred imagery.

\begin{figure}[t]
  \centering
  \subfigure[Artful motion blur.]{
  \label{fig:motion_blur2}
  \includegraphics[width=0.47\columnwidth]{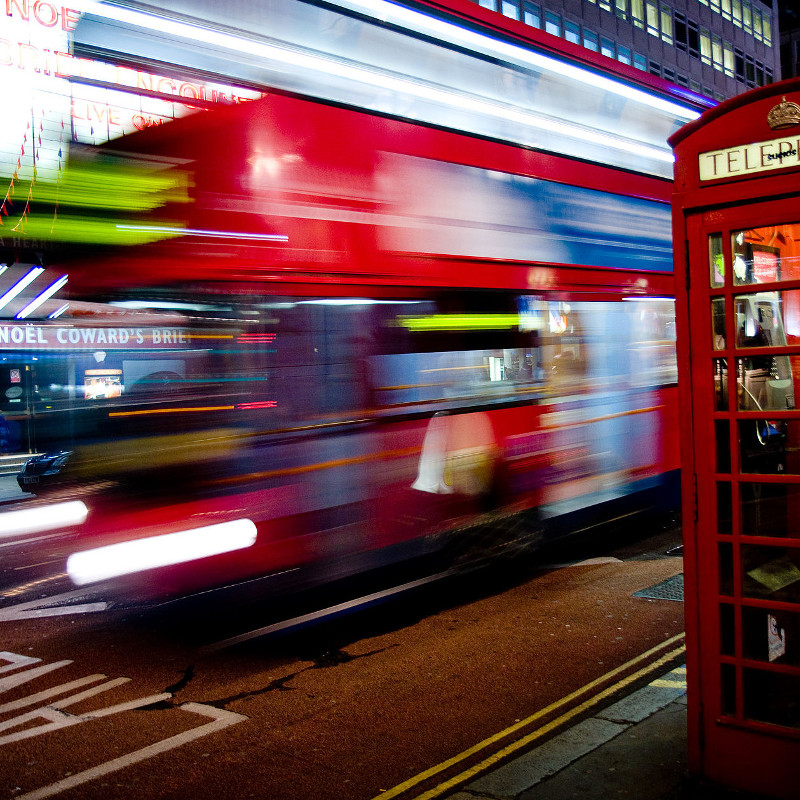}
  }
  \subfigure[Unwanted motion blur.]{
  \label{fig:motion_blur1}
  \includegraphics[width=0.47\columnwidth]{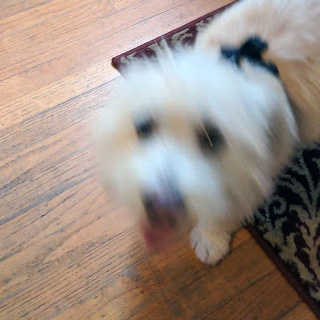}
  }
  \caption{
  Capable photographers can use motion blur to produce striking photographs, as in \subref{fig:motion_blur2}.
  But for most casual photographers, motion blur is more likely to manifest as an unwanted artifact in an image that was intended to be completely sharp, as in \subref{fig:motion_blur1}.
  \label{fig:motion_blur}
  }
\end{figure}

Talented photographers sometimes use motion blur for artistic effect (Figure~\ref{fig:motion_blur2}). But composing an artful motion-blurred photograph is a difficult process, typically requiring a tripod, manual camera settings, perfect timing, expert skill, and much trial and error.
As a result, for casual photographers motion blur is likely to manifest as an unwanted artifact (Figure~\ref{fig:motion_blur1}).
Because of the difficulty in using motion blur effectively, most consumer cameras are designed to take images with as little motion blur as possible --- though if noise is a concern \emph{some} motion blur is unavoidable, especially in low-light environments or in scenes with significant motion~\cite{Hasinoff2016}.
Artistic control over motion blur is therefore out of reach for most casual photographers.
By allowing motion blurred images to be synthesized from the conventional unblurred images that are captured by standard consumer cameras, our technique allows non-experts to create motion blurred images in a post-capture setting.
This is analogous to how recent progress in depth estimation has enabled post-capture on-device depth-of-field manipulation, also known as ``Portrait Mode'' \cite{ApplePortraitMode,Barron2015A,portraitmode2018}.

Motion blur is also an important tool in cinematography, where filmmakers will carefully adjust the shutter angle of their camera to create a particular ``film look''. As in photography, this requires expert domain knowledge and skillful execution.
Our system (or indeed any system that operates on pairs of frames) can be used to manipulate the motion blur of video sequences after the fact, by independently processing all pairs of adjacent frames in the input video.

Motion blur synthesis has been extensively studied in the rendering community \cite{MotionBlurRendering}, though these methods typically require perfect knowledge of scene velocities and depths as inputs. We instead target the most general form of this problem, and assume the only inputs available to our system are unblurred input images, as is the case in most general vision and imaging contexts.

To enable the varied image understanding and image manipulation tasks that require a method for creating motion blur, we present an algorithm that takes two sharp images taken one after the other, as shown in Figure~\ref{fig:teaser-a}, and synthesizes a corresponding motion blurred image, such as in Figure~\ref{fig:teaser-b}. The synthesized image resembles an image captured over the time spanned by the input images --- the image ``starts'' at the first input image, and ``ends'' at the second input image. To achieve this, we adapt recent advances in machine learning to the task of predicting line kernels for motion blurring image pairs. 

We build upon the recent success of convolutional neural networks \cite{Lecun1989} and end-to-end training on tasks similar to ours, such as optical flow \cite{Flownet2,Sun2018PWCNet,xue17toflow} and frame interpolation \cite{SuperSlomo2018,NiklausCVPR2018,NiklausCVPR2017,NiklausICCV2017}. We use state-of-the-art frame interpolation to synthesize training data for our motion blur model, and demonstrate that our model, trained directly on the task of synthesizing motion blur, produces improved results on real images over baselines derived from optical flow and frame interpolation techniques. Though frame interpolation achieves closer accuracy, our technique is significantly faster, and is thereby better suited to the online synthesis of training data in a deep learning context, and is easier to deploy in a consumer-facing rendering or smartphone-photography setting.

The remainder of this paper is structured as follows: In Section~\ref{sec:problem} we discuss the nature of motion blur as a function of linear motion and motivate our novel line prediction layer. In Section~\ref{sec:arch} we define a deep neural network architecture based on our line prediction layer. In Section~\ref{sec:dataset} we construct a synthetic dataset that is used for training, and a real-world dataset that is used for evaluation. In Section~\ref{sec:experiments} we evaluate the performance of our model compared to its ablations and variants, and to techniques in the literature that can be adapted to the task of synthesizing motion blur.

\section{Problem Formulation}
\label{sec:problem}

We aim to take two adjacent images from a camera, say from a video or from a ``burst'' of photos~\cite{Hasinoff2016}, and from them synthesize a motion blurred image that spans the duration between the input images. That is, letting $I_1$ be the image exposed for the duration $[s_1, t_1]$ and $I_2$ be the image exposed for the duration $[s_2, t_2]$ (where $s_1 < t_1 < s_2 < t_2$), we synthesize the long exposure photograph $I_{1 \rightarrow 2}$, which spans the duration $[s_1, t_2]$.

Similar to the assumptions of optical flow, which describes motion between two frames in terms of per-pixel velocity vectors, we assume locally linear motion between the two input images. We further assume that each pixel in the motion blurred image can be linearly interpolated from pixels lying on lines drawn from the corresponding pixel in each of the input images. While these assumptions are not always valid---for example, in the case of objects that are rotating or oscillating---we will demonstrate that this simple linear model is sufficiently expressive to produce high quality results.

Our neural network architecture uses a novel ``line prediction'' layer, which we define here. For each pixel in our images $I_i$ ($i \in \{1, 2\}$) we predict a line, where one endpoint of that line is at the pixel's location $(x, y)$ and the other endpoint is at $(x + \Delta_i^{x}(x, y), y + \Delta_i^{y}(x, y))$---the pixel's location when advected by some predicted offset $\Delta_i$. The line is composed of $N$ evenly-spaced discrete samples, for which we also predict $W_i(x, y, n)$, a weighting for each sample.
Our final predicted image $I_{1 \rightarrow 2}$ is defined as the weighted average of the two input images according to the discrete samples along all lines:
\begin{equation}
    I_{1 \rightarrow 2}(x, y) =  \sum_{i \in \{1, 2\}} \sum_{n=0}^{N-1} W_i( x, y, n) \times \quad\quad \label{eq:line}
\end{equation}
\vspace{-0.1in}
\begin{equation}
    I_i\left(x + \left(\frac{n}{N-1}\right) \Delta_i^{x}(x, y), y + \left(\frac{n}{N-1}\right) \Delta_i^{y}(x, y) \right), \nonumber
\end{equation}
where $I_i(x, y)$ is the result of bilinear interpolation of $I_i$ at any continuous location $(x, y)$.

We refer to this approach as ``line prediction'', analogously to the ``kernel prediction'' literature \cite{MonteCarloKPN,Mildenhall2018,NiklausCVPR2017}. Our model can be thought of as a form of kernel prediction, as the weighted average in Equation~\ref{eq:line} can be rasterized into a per-pixel convolution with a discrete kernel composed of the sum of the weighted bilinear interpolation kernels used in line prediction---though reformulating the blur in this way makes it significantly more expensive to compute.

For our line prediction technique to work properly, we must reason about the relationship between our line offsets $\Delta_i$ and our sampling density. Since the standard deep learning techniques we use for estimating the parameters of our line prediction layer have difficulty producing variable-length outputs, the number of estimated line samples $N$ is fixed. However, if the motion estimated at a given pixel is significantly greater than the number of samples available to reconstruct our predicted line, then our resulting motion blurred image will be \emph{temporally undersampled}, and will therefore contain artifacts from these  ``gaps'' when synthesizing motion blur. See Figure~\ref{fig:sampling} for a visualization of this sampling issue. For this reason, when determining a value for $N$, we must impose a bound on the magnitude of our line endpoint displacements $(\Delta_i^{x}(x, y), \Delta_i^{y}(x, y))$. We only address the task of synthesizing motion blurred images whose maximal displacement is $32$ pixels in length, and we set $N=17$. We found that we are able to use half as many samples as our maximum displacement because the kernel used by bilinear interpolation effectively prefilters the convolution induced by our line prediction. This limit on pixel displacement and sampling density is analogous to the similar limits of kernel prediction-based video frame interpolation techniques with regard to their kernel sizes.

\newcommand{\halfwidth}{0.47\linewidth}
\begin{figure}  \centering
  \subfigure[Temporal undersampling]{
  \label{fig:sampling1}
  \includegraphics[width=\halfwidth]{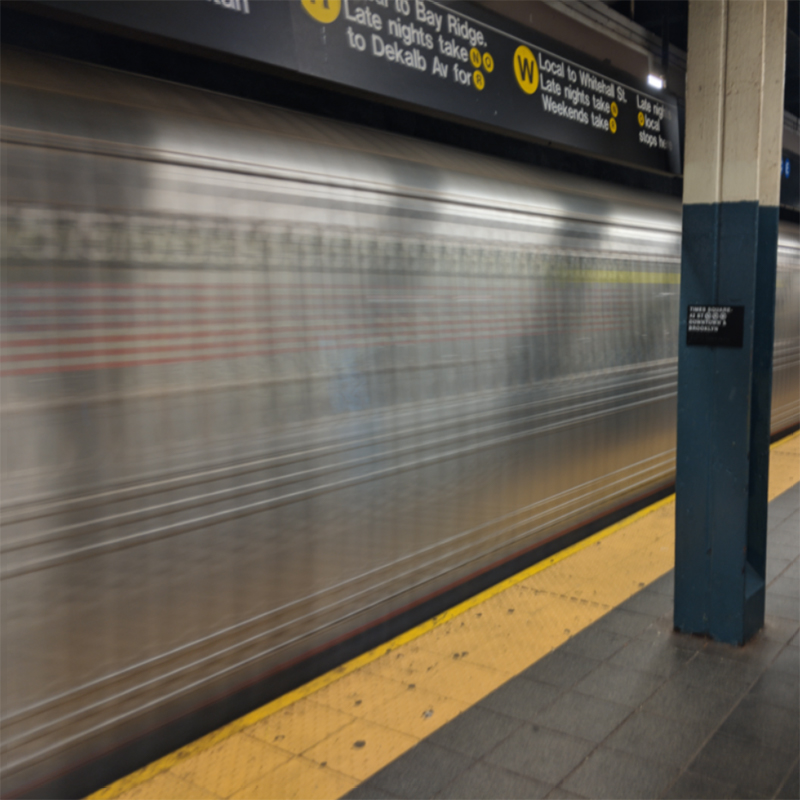}
  }
  \subfigure[Temporal supersampling]{
  \label{fig:sampling2}
  \includegraphics[width=\halfwidth]{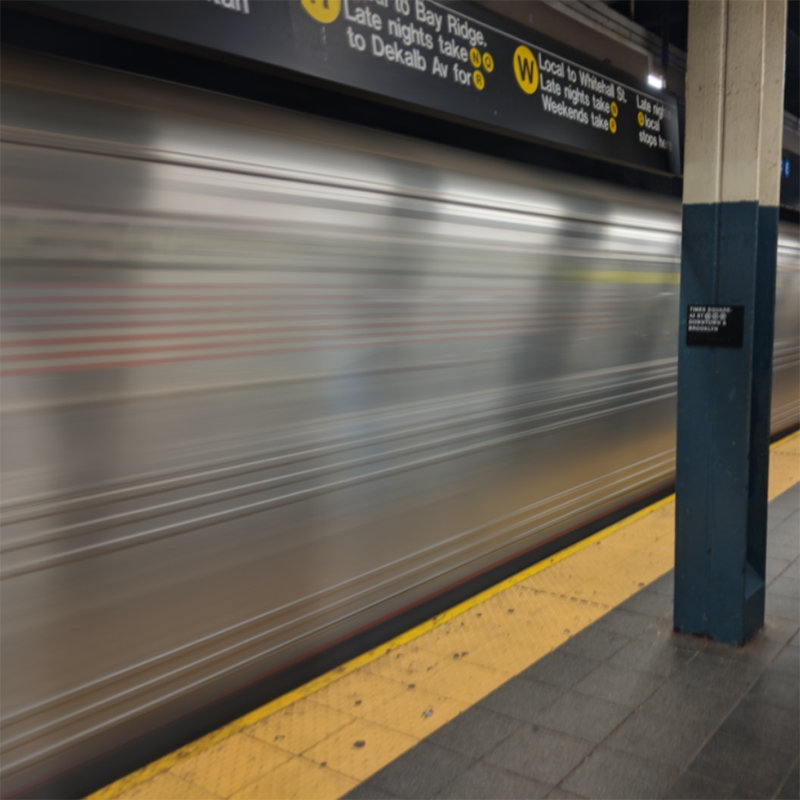}
  }
  \caption{Temporal sampling is critical to the construction of our model and our training data. If a motion blurred image is synthesized using significantly fewer samples than the maximum displacement of any pixel across those samples, then that synthesized image may be temporally undersampled. This results in discontinuous artifacts along the direction of the motion, as in \subref{fig:sampling1}. If the sampling density is sufficiently large with respect to image resolution and object motion then the synthesized images will not exhibit any such artifacts, as in \subref{fig:sampling2}.
  \label{fig:sampling}
  }
\end{figure}

\begin{figure*}
  \centering
  \includegraphics[width=0.8\linewidth]{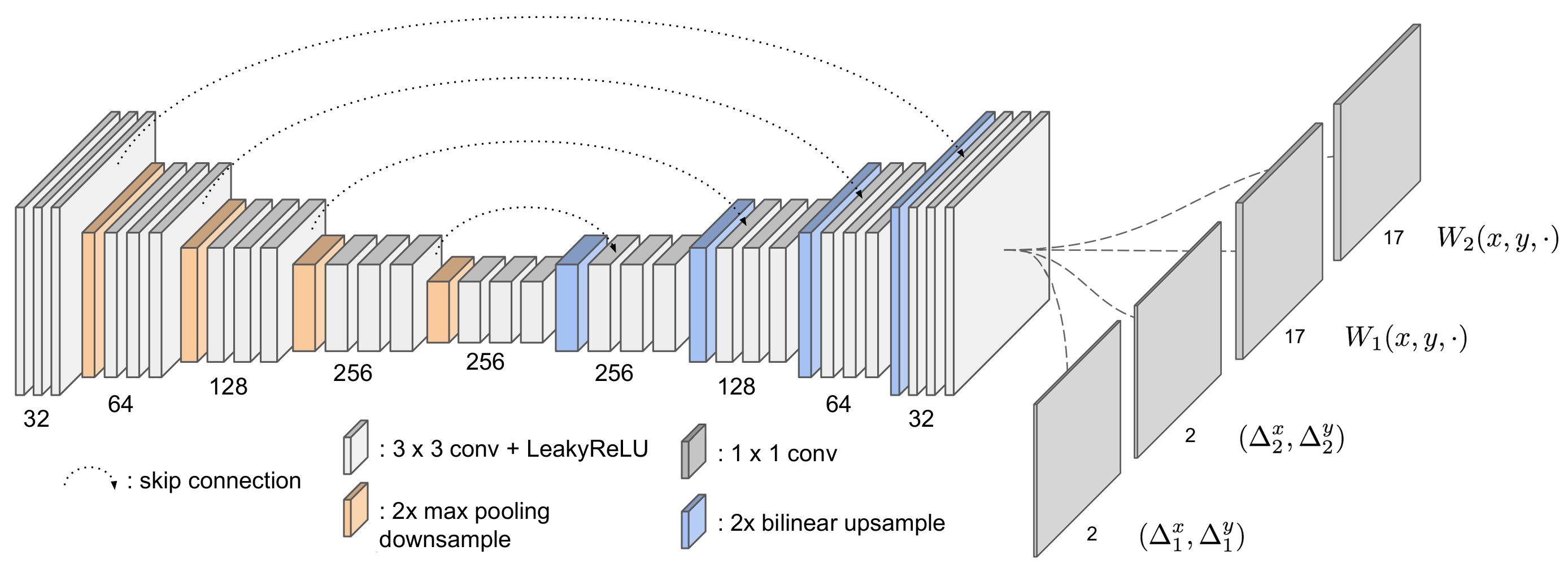}
  \caption{
  A visualization of our architecture, which takes as input a concatenation of our two input images and uses a U-Net convolutional neural network to predict the parameters for our line prediction layer.
  \label{fig:arch}
  }
\end{figure*}

Our decision to have our network predict a set of sampling weights $W_i(x, y, n)$ may seem unusual, as techniques from the graphics literature tend to assign uniform weights to pixels when rendering motion blur \cite{PlausibleMotionBlur}. These learned weights allow our algorithm to handle complex motions and occlusions, and to hedge against certain failure modes. For example, by emitting a weight of $0$, our model can ignore certain pixels during integration, which may be necessary if the pixel of interest moves behind an occluder on its path towards its location in the other frame. Because our synthesis happens simultaneously in both the ``forward'' and ``backward'' direction, our model can use these weights to smoothly transition across images or to selectively draw from one image but not the other, further improving its ability to reason about occlusion. Though our model is constrained to linear motion, these weights can be used to model an object as moving at a non-constant speed along its line. For example, if an object accelerates towards its destination, our model can synthesize a more accurate motion blur (without introducing any temporal undersampling issues) by giving early samples higher weights than later samples.

\section{Model Architecture}
\label{sec:arch}

Our model is built around the U-Net architecture of \cite{U-Net}, which feeds into our line prediction layer whose output is used to synthesize a motion blurred image. The input to our model is simply the concatenation of our two input images. See Figure~\ref{fig:arch} for a visualization of our architecture.

The U-Net architecture, which has been used successfully for the related task of frame interpolation \cite{SuperSlomo2018,NiklausICCV2017}, is a fully-convolutional encoder/decoder model with skip connections from each encoder to its corresponding decoder of the same spatial resolution. Our encoder consists of five hierarchies (sets of layers operating at the same scale) each containing three `conv' layers, and where all but the last hierarchy are followed by a max pooling layer that downsamples the spatial resolution by a factor of 2$\times$. Our decoder consists of four hierarchies, each with three conv layers that are followed by a bilinear upsampling layer that increases spatial resolution by a factor of 2$\times$. Each conv layer uses 3$\times$3 kernels and is followed by leaky ReLU activation~\cite{LeakyRelu}.

We train our model end-to-end by minimizing the L1 loss between our model's predicted motion blurred image and our ground-truth motion blurred images.
Our data augmentation and training procedure will be described in more detail in Section~\ref{sec:experiments}.
We experimented with pretraining our line prediction model using optical flow training data, as prescribed in \cite{xue17toflow}, but this did not appear to improve performance or significantly speed up convergence.
Our model is implemented using TensorFlow~\cite{Abadi2016}.

\section{Dataset}
\label{sec:dataset}

Training or evaluating our model requires that we produce ground truth data of the following form: two input images, and an output image wherein the camera has integrated light from the start of the first image to the end of the second image.
Because large neural networks require an abundance of data, for training we present our own synthetic data generation technique based around video frame interpolation, which we use to synthesize motion blurred images from conventional, abundantly available video sequences (Sec~\ref{sec:synth_data}). We take sets of adjacent video frames, synthesizing many intermediate images between those frames, and average all resulting frames to make a single synthetic motion blurred image (where the original two frames can then be used as input to our algorithm). These synthesized motion blurred images look reasonable and are easy to generate in large quantities, but they may contain artifacts due to mistakes in the underlying video frame interpolation technique and so have questionable value as a ``test set''. Therefore, for evaluation, where data fidelity is valued more highly than quantity, we use a small number of real slow-motion video sequences. The first and last frames of each sequence are used as input to our algorithm, and the sum of all frames in the sequence is used as the ``ground-truth'' motion blurred image (Sec~\ref{sec:realdata}).

\subsection{Synthetic Training Data}
\label{sec:synth_data}

We manually created our own dataset directly from publicly available videos, as this gives us precise control over things like downsampling and the amount of motion present in the scene, while allowing us to select for interesting, high-frequency scene content. To construct this dataset, we first extract sets of adjacent triplets from carefully chosen video sequences, and then use those triplets to train a video frame interpolation algorithm. This video frame interpolation algorithm is then applied recursively to all triplets, which allows us to synthesize a $33$ frame interpolated sequence from each triplet that can then be averaged to produce a synthetically motion blurred image. These images are then treated as ``ground truth'' when training our model.

We downloaded $\sim$30,000 Creative Commons licensed 1080p videos from YouTube in categories that tend to have significant amounts of motion, such as ``Wildlife,'' ``Extreme Sports,'' and ``Performing Arts.'' We then downsampled each video by a factor of 4$\times$ using bicubic interpolation to remove compression artifacts, and then center-cropped each sequence to a resolution of 270$\times$270. From these video sequences, we extracted triplets of adjacent frames that satisfy the following properties:
\begin{enumerate}[noitemsep,nolistsep,leftmargin=0pt, itemindent=15pt]
    \item {\bf High frequency image content}: Focusing training on images with interesting gradient information tends to improve training for image synthesis tasks such as our own, as shown in \cite{Gharbi2016}. We therefore rejected any triplet whose average gradient magnitude (computed using Sobel filters) over all pixels was less than $13$ (assuming images are in $[0, 255]$).
    \item {\bf Sufficient motion}: Scenes without motion are unlikely to provide much signal during training. Therefore, for each triplet we estimated per-pixel motion across adjacent frames (using the fast optical flow technique of \cite{celiu}) and only accepted triplets where at least $10\%$ of each pixel's flow had a magnitude ($\infty$-norm) of at least $8$ pixels.
    \item {\bf Limited motion}: Our learned model and many of the baseline models we compare against have outputs with limited spatial support, and we would like our training data to lie entirely within the receptive field of our models. We therefore discarded any triplet that contained a flow estimate with a magnitude ($\infty$-norm) of more than $16$.
    \item {\bf No abrupt changes}: Significant and rapid changes across adjacent frames in our video data are often due to cuts or other kinds of video editing, or global changes in brightness or illumination. To address this, we warp each frame in each triplet according to its estimated motion and discard triplets with an average L1 distance of more than $13$ (assuming images are in $[0, 255]$).
    \item {\bf Approximately linear motion}: Our model architecture is only capable of estimating and applying a linear motion blur. Images that are not expressible using linear blurs will therefore likely not contribute much signal during training. We therefore compare the ``forward'' flow between the second and third frame to the negative of the ``backward'' flow from the first and second frame, and discard any triplets with a mean disagreement of $>$0.8 pixel widths.
\end{enumerate}

\newcommand{\quarterwidth}{0.245\linewidth}
\begin{figure}
  \centering
  
  \begin{tabular}{@{}c@{}c@{}c@{\,\,}c@{}}
  \includegraphics[width=\quarterwidth]{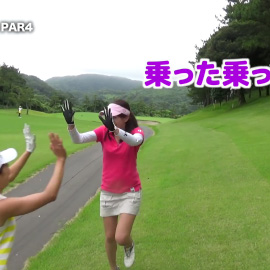}
  &
  \includegraphics[width=\quarterwidth]{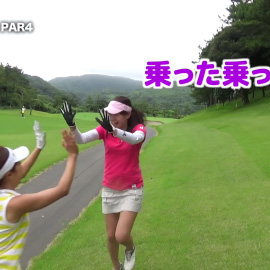}
  &
  \includegraphics[width=\quarterwidth]{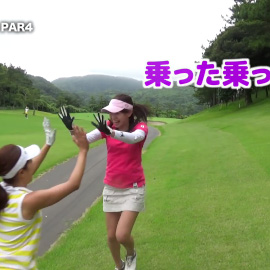}
  &
  \includegraphics[width=\quarterwidth]{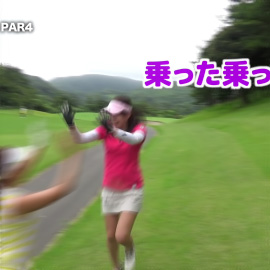}
  \\
  \includegraphics[width=\quarterwidth]{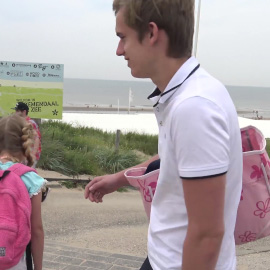}
  &
  \includegraphics[width=\quarterwidth]{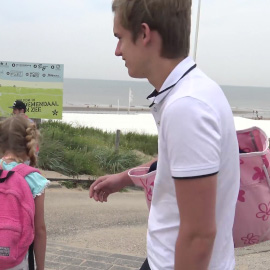}
  &
  \includegraphics[width=\quarterwidth]{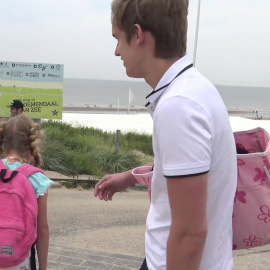}
  &
  \includegraphics[width=\quarterwidth]{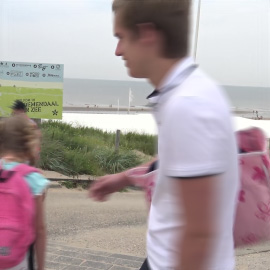}
  \\
  \includegraphics[width=\quarterwidth]{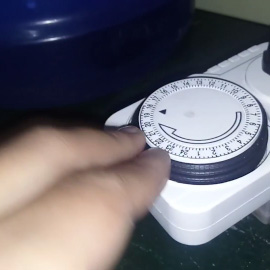}
  &
  \includegraphics[width=\quarterwidth]{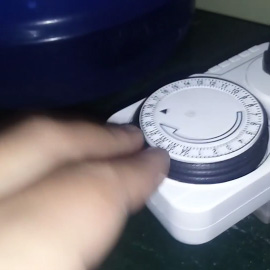}
  &
  \includegraphics[width=\quarterwidth]{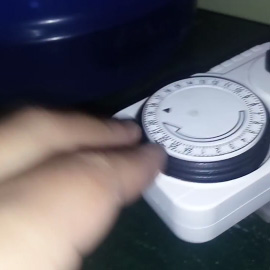}
  &
  \includegraphics[width=\quarterwidth]{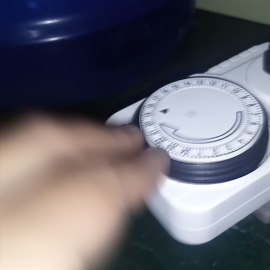}
  \\
  \includegraphics[width=\quarterwidth]{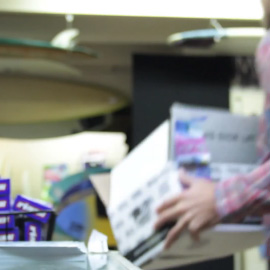}
  &
  \includegraphics[width=\quarterwidth]{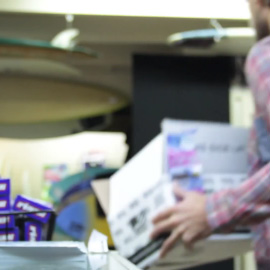}
  &
  \includegraphics[width=\quarterwidth]{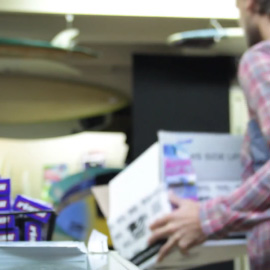}
  &
  \includegraphics[width=\quarterwidth]{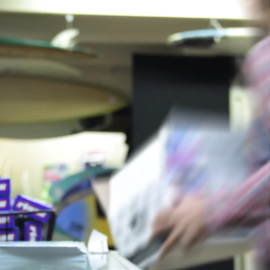}
  \\
  \end{tabular}
  \caption{Here we show randomly chosen input/output pairs from our synthetic training dataset. To generate this data we identify triplets (shown in the first three columns) of adjacent frames that satisfy our criteria for motion and image content, use those triplets to train a video frame interpolation model, and apply that model recursively on each triplet to generate intermediate frames which are then averaged to synthesize a single motion blurred image (shown in the last column). When training our motion blur model, we use the first and last images of each triplet as input and the averaged image as ground-truth.
  \label{fig:training_data}
  }
\end{figure}

Note that (5) represents a kind of ``co-design'' of our algorithm and our training data, in that we craft our dataset to complement the assumptions of our model. To evaluate a broader generalization of our model, we do not impose this constraint on our ``real'' testing dataset.

To ensure diversity, we extract no more than $50$ triplets from each video, and no more than a single triplet from a given scene within each video. This process resulted in $>$300,000 unique triplets, of which $5\%$ are set aside for validation with the remaining $95\%$ used for training. This training/validation split is carefully constructed such that all triplets generated from any given video are assigned to either the training or validation split --- no video's triplets are present in both the training or validation splits.

With this dataset we then train a video frame interpolation network based on \cite{NiklausICCV2017}, which will shortly be used to produce the final motion blur training data we are pursuing. Our frame interpolation network is the same model as described in Section~\ref{sec:arch}, but using a separable kernel prediction layer of 33$\times$33 learned kernels instead of our line prediction layer. Our training procedure is described in more detail in Section~\ref{sec:experiments}. The need to train this frame interpolation model is why we chose to extract triplets from our video sequences as opposed to just two frames, as the middle frame of each triplet can be used as ground-truth during this training stage (but will be ignored when training our motion blur model).
After training, this frame interpolation model takes two frames as input, and from them synthesizes an output frame that should lie exactly in between the two input frames. We apply this network to our triplet of video frames, first using the first and second frames of the triplet as input to the network to synthesize an in-between frame, then using the second and third frames to synthesize another in-between frame. We then apply this same process recursively using the real and newly-interpolated frames as input. This is done $4$ times, resulting in a $33$ frame sequence of interpolated frames. These frames are all then averaged to produce a synthetically motion blurred image. Note that our recursive interpolation process yields $15$ frames between each image in our triplet. Because our previously-described data collection procedure omitted adjacent frames with a motion of more than $16$ pixels, this means that we should expect our interpolated images to have a motion of less than one pixel width per frame. This means that our resulting motion blurred images should not suffer from temporal undersampling. See Figure~\ref{fig:training_data} for some examples of our synthetic training data.

\subsection{Real Test Data}
\label{sec:realdata}

For evaluation purposes we would like a small, high-quality dataset that is not vulnerable to the artifacts that may be introduced by frame interpolation algorithms, and is as close as possible to a real in-camera motion blurred image.
Although it is easy to acquire motion blurred images by themselves, acquiring the two input images alongside that motion blurred image is not possible with conventional camera sensors.
We therefore capture a series of short slow motion videos, where the first and last frames of each video are used as input to our system, and the per-pixel mean of all frames is used as the ground-truth motion blurred image.
Our dataset was gathered by a photographer using the Panasonic LUMIX GH5s, which records videos at 240fps. The photographer was instructed to photograph subjects that are well-suited to an artistic use of motion blur: people walking or running, vehicles moving, falling water, etc. Images were bicubicly downsampled by 2$\times$ to help remove demosaicing and compression artifacts, and center-cropped to 512$\times$512 pixels.
From each video we selected a span of frames such that the total motion across the span is no more than $32$ pixels.
Any sequences that exhibited any temporal undersampling were removed.
For each sequence we generated a single motion blurred image by simply averaging the frames, and we set aside the first and last frame of each sequence for use as input to our model.
Each sequence has a variable length of frames, as we saw no need to omit frames from each sequence if they happened to be temporally super-sampled. Our final dataset consists of $21$ diverse sequences. See Figure~\ref{fig:results1} and the appendix for examples.

\section{Experiments}
\label{sec:experiments}

\newcommand{\resultswidth}{0.23\linewidth}
\newcommand{\resultname}{16_basketball}
\begin{figure*}[t!]
  \centering
  \subfigure[Input image 1]{
  \label{fig:results11}
  \includegraphics[width=\resultswidth]{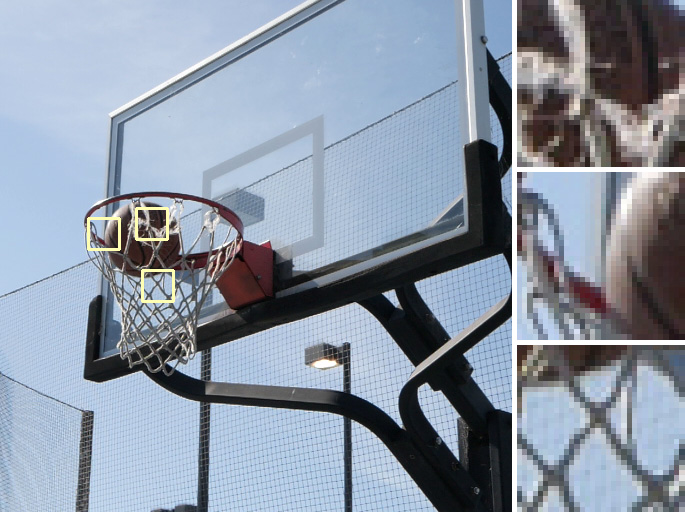}
  }
  \subfigure[Input image 2]{
  \label{fig:results12}
  \includegraphics[width=\resultswidth]{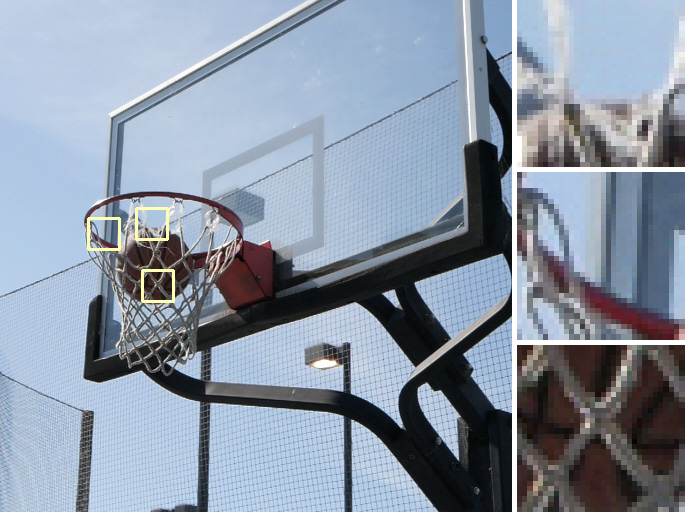}
  }
  \subfigure[Non-input intermediate frames]{
  \label{fig:results13}
  \includegraphics[width=\resultswidth]{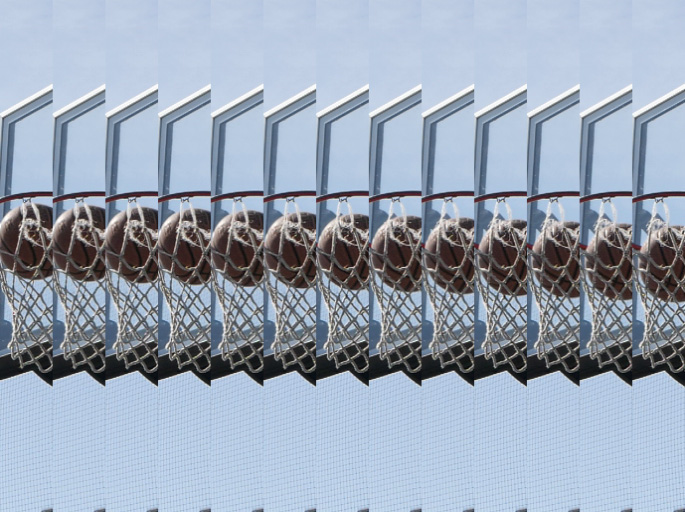}
  }
  \subfigure[Ground-truth motion blur]{
  \label{fig:results14}
  \includegraphics[width=\resultswidth]{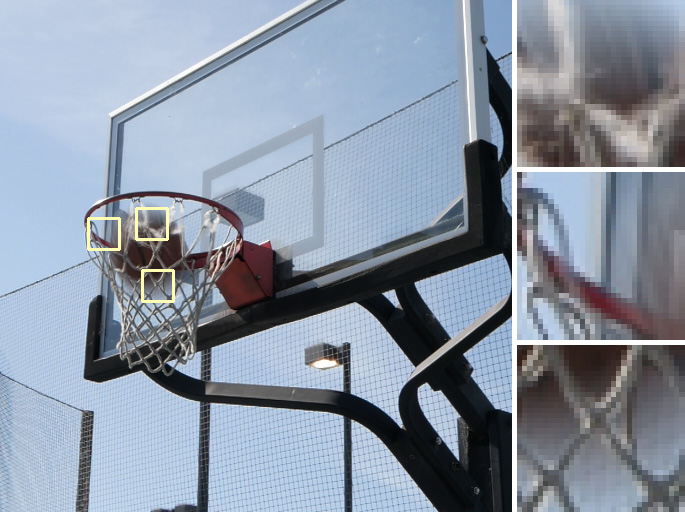}
  }
  \subfigure[\namepwc]{
  \includegraphics[width=\resultswidth]{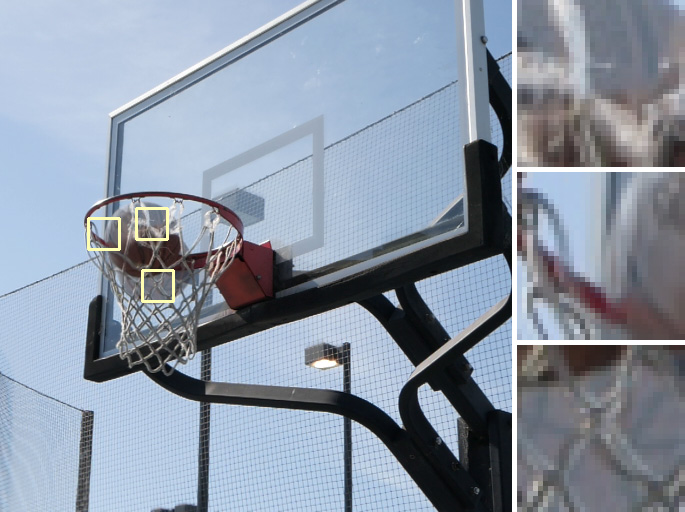}
  \label{fig:results15}
  }
  \subfigure[\nameepic]{
  \includegraphics[width=\resultswidth]{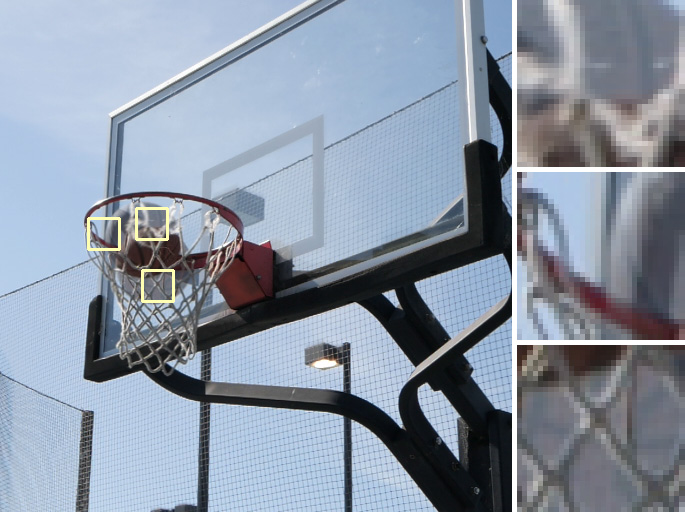}
  \label{fig:results16}
  }
  \subfigure[\namesepconv]{
  \includegraphics[width=\resultswidth]{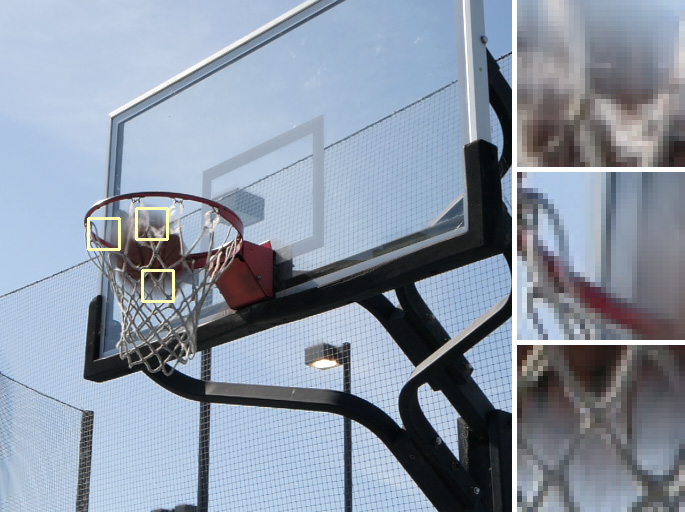}
  \label{fig:results17}
  }
  \subfigure[\nameslomo]{
  \includegraphics[width=\resultswidth]{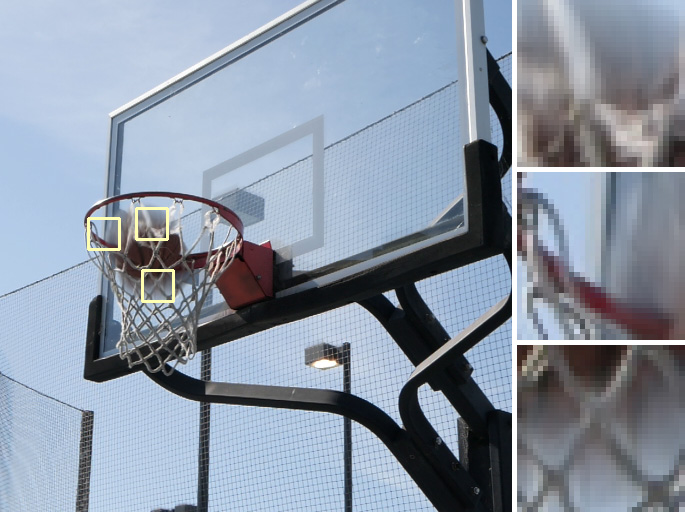}
  \label{fig:results18}
  }
  \subfigure[\namedirect]{
  \includegraphics[width=\resultswidth]{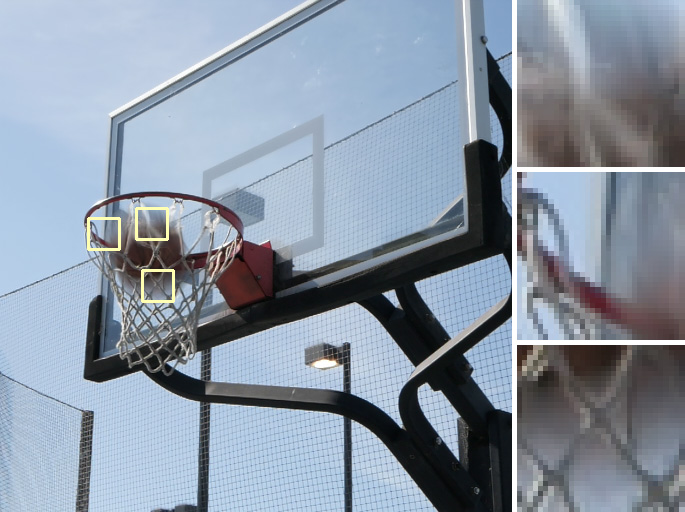}
  \label{fig:results19}
  }
  \subfigure[\nameavg]{
  \includegraphics[width=\resultswidth]{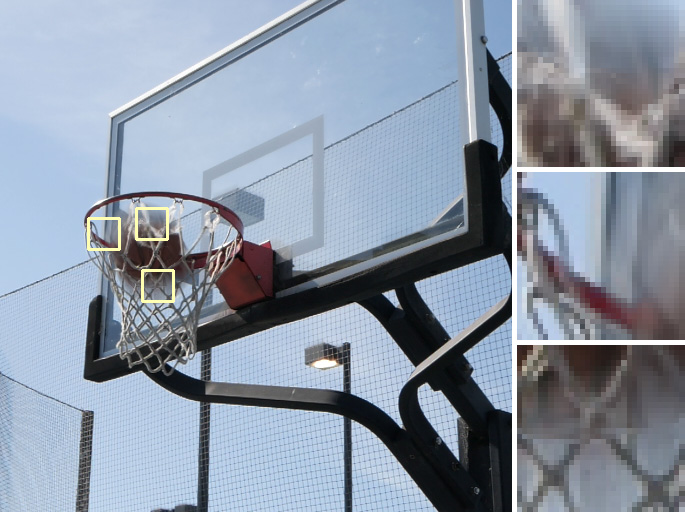}
  \label{fig:results19a}
  }
  \subfigure[\namekpn]{
  \includegraphics[width=\resultswidth]{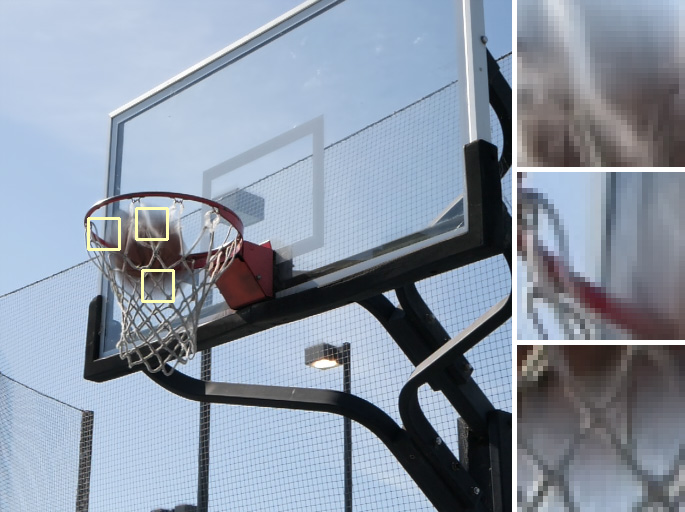}
  \label{fig:results19b}
  }
  \subfigure[\namemodel]{
  \includegraphics[width=\resultswidth]{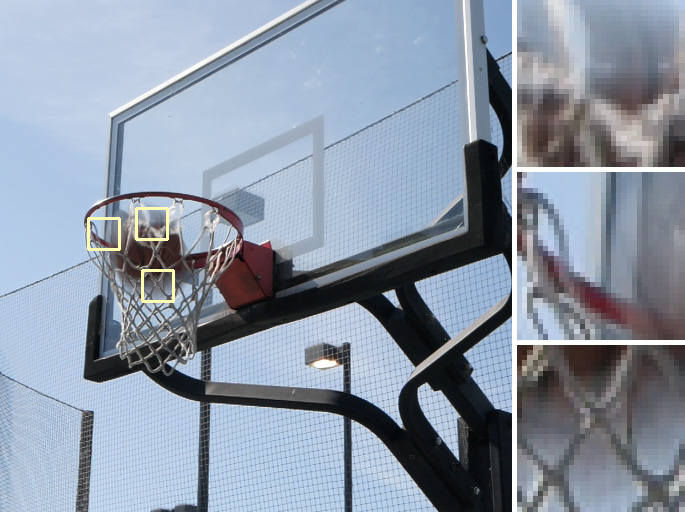}
  \label{fig:results19c}
  }
  \caption{
  Results for one scene from our test dataset.
  The ground truth image \subref{fig:results14} is the sum of the input images \subref{fig:results11} \& \subref{fig:results12} and of the frames between those two images \subref{fig:results13}.
  We programmatically select the three non-overlapping 32$\times$32 sub-images with maximal variance across all frames in \subref{fig:results13} and present crops of those regions, rendered with nearest-neighbor interpolation and sorted by their $y$-coordinates. We compare our model \subref{fig:results19c} against four baselines \subref{fig:results15}-\subref{fig:results18}, and three ablations \subref{fig:results19}-\subref{fig:results19b}.
  See the appendix for additional results.
  \label{fig:results1}
  }
\end{figure*}

Our motion blur models, as well as our frame interpolation model used to generate our synthetic data, were trained distributedly over $8$ NVIDIA Tesla P100 GPUs for $3.5M$ iterations on batches of size $16$ using the Adam optimization algorithm \cite{AdamOpt} with a learning rate of $\alpha = 0.00002$ and momentum decay rates $\beta_1 = 0.9$ and $\beta_2 = 0.998$.
During training we performed data augmentation by randomly extracting a 256$\times$256 crop from each image, and then randomly applying a horizontal flip, vertical flip, and a $90^\circ$ rotation.
Training to convergence took $\sim$2.5 days.

We  evaluate our model against five baseline algorithms:
A ``naive'' baseline that is simply the mean of the two input images (see Figure~\ref{fig:flow1}),
the non-learned and non-deep optical flow algorithm of ~\cite{EpicFlow},
the state-of-the-art learned flow method of \cite{Sun2018PWCNet},
the video frame interpolation work of \cite{NiklausICCV2017} (which improves upon \cite{NiklausCVPR2017}),
and the state-of-the-art video interpolation work of \cite{SuperSlomo2018}.
We additionally evaluate against three ablated versions of our model:
\begin{enumerate}[noitemsep,nolistsep,leftmargin=0pt, itemindent=15pt]
    \item {\bf Direct Prediction}: instead of using line prediction our network directly estimates the motion blurred image, by replacing our line prediction model with a single 1$\times$1 conv layer that produces a $3$ channel output.
    \item {\bf Uniform Weight}: we use uniform weights for each sample along lines rather than learning weights (i.e., all $W_i(x, y, n) = \sfrac{1}{2N}$).
    \item {\bf Kernel Prediction}: instead of using line prediction we use the separable kernel prediction of \cite{NiklausICCV2017}, by replacing our line prediction layer with a single 1$\times$1 conv layer at the end of our network that produces a 65$\times$65 separable kernel (represented as a 65$\times$1 and 1$\times$65 kernel) at each pixel.
\end{enumerate}

Our ``kernel prediction'' model has an inherent limitation, as separable kernels are limited in their ability to represent angled blur kernels. For example, the matrix corresponding to a blur kernel of a diagonal line is full-rank and cannot be represented well as a rank-1 matrix, so equivalently, the kernel cannot be represented well by a separable kernel. This limitation can be addressed by using non-separable kernels as in \cite{NiklausCVPR2017}, however, the large kernels needed for our application require extreme amounts of memory that far exceeded the limits of our GPUs when we attempted to use this approach for training.

To generate motion blurred comparisons from our optical flow baselines, we employed the same line blurring scheme as our ''uniform weight'' model, and bilinearly sample $N$ evenly-spaced values from the input images along lines corresponding to the optical flow fields. These sampled images are then averaged to produce a motion blurred image.
We found that both flow algorithms benefited significantly (a PSNR improvement of $\sim$5) from using the negative backward flow instead of the forward flow to produce motion blur, so we adopted that strategy when evaluating our baseline flow techniques.
More sophisticated strategies for gathering and scattering in forward and backward directions of object velocities have been used to synthesize motion blur in the graphics literature  \cite{PlausibleMotionBlur,MotionBlurRendering}, but these techniques assume that perfect scene geometry is known and so cannot be used for our task.

Comparisons against frame interpolation baselines were conducted by recursively running frame interpolation on the input image pair for $5$ iterations, which results in a $33$-frame sequence --- a sufficiently dense sampling given the limit of $32$-pixel displacements in our real test set. The resulting synthetic slow motion sequences were then averaged to produce a motion blurred image.

\definecolor{Gray}{rgb}{0.85, 0.85, 0.85}
\definecolor{Yellow}{rgb}{1, 1, 0.7}
\definecolor{Orange}{rgb}{1, 0.7, 0.7}

\begin{table}[b]
\centering
\Large
\resizebox{\linewidth}{!}{%
\begin{tabular}{l|ccc}
Algorithm              & PSNR & SSIM & Runtime (ms) \\
\hline
\namebaseline  &                      $28.06 \pm 4.05$ &                      $0.888 \pm 0.087$ & - \\
\namepwc &                      $29.93 \pm 3.47$ &                      $0.938 \pm 0.057$ & $39.5$ \\
\nameepic  &                      $30.07 \pm 3.49$ &                      $0.940 \pm 0.057$ & $96.3 \times 10^6$ \\
\namesepconv &                      $32.91 \pm 4.60$ &                      $0.954 \pm 0.054$ & $10.9 \times 10^4$ \\
\nameslomo &                      $33.64 \pm 4.66$ &                      $0.958 \pm 0.048$ & $13.7 \times 10^2$ \\
\hline 
\namedirect &   \cellcolor{Yellow} $33.97 \pm 4.53$ &                      $0.961 \pm 0.044$ & $34.7$ \\
\nameavg &                      $33.88 \pm 4.68$ &                      $0.959 \pm 0.050$ & $42.8$ \\
\namekpn &                      $33.73 \pm 4.31$ &   \cellcolor{Yellow} $0.961 \pm 0.045$ & $65.5$ \\
\namemodel &   \cellcolor{Orange} $34.14 \pm 4.65$ &   \cellcolor{Orange} $0.963 \pm 0.045$ & $43.7$ \\
\end{tabular}
}
\caption{Performance on our real test dataset, in which we compare our model to three of its ablated variants and five baseline algorithms.
}
\label{table:real_results}
\end{table}

We primarily evaluate our model on the real test dataset described in Section~\ref{sec:realdata}, shown in Table~\ref{table:real_results}. We report the mean PSNR and SSIM for the dataset, and note that our model produces the highest value of both out of all baselines and ablations. Though at first glance the difference between models may appear small, the unusually high PSNR of the ``naive'' baseline serves to anchor these scores and suggests that small variations in scores are meaningful.
The two optical flow baselines are the lowest-performing techniques, with the two video frame interpolation techniques performing nearly as well as ours. However, the gap in runtime between our model and the baseline techniques is quite substantial, as our model is 30$\times$ - 2,500$\times$ faster. This is partially due to our compact architecture and the fact that line prediction is amenable to a fast implementation, but is also because video frame interpolation techniques must predict a $33$ frame sequence that is then averaged to produce a single image, and so necessarily suffer a 33$\times$ speed decrease.

The reported runtimes of our model, its ablations, and the technique of \namesepconv\, are the mean of 1000 runs on a GeForce GTX 1080 Ti, at our test set image resolution of 512$\times$512. The runtimes of \namepwc\, and \nameslomo\, were reported by the authors of those papers, who graciously ran their code on our data using a NVIDIA Pascal TitanX (a faster GPU than the one used for our model). The runtime for \nameepic\, was extrapolated from the numbers cited in the paper, which were produced on a 3.6Ghz CPU. Reported times for the optical flow methods are underestimates of their true runtimes, as we only measure the time taken to generate their flow fields, and do not include the time taken to render images from those flow fields.

The reduced performance of our ``uniform weight'' ablation appears to be due to its difficulty in handling occlusions and motion boundaries, which appear to particularly benefit from the learned sample weights. This can be seen in Figure~\ref{fig:results19c}, where our model appears to use its learned weights to blur around the occlusions of the basketball net webbing.

The output of our model superficially resembles an optical flow algorithm, in that the line endpoint $\Delta_i^{x}(x, y)$ predicted at each pixel can be treated as a flow vector. Though this is an oversimplification (our model actually predicts a weighting for a set of points along this line and those weights may be zero, effectively shortening or shifting this line) it is illustrative to visualize our output as a flow field and compare it to optical flow algorithms, as in Figure~\ref{fig:flowvis}. Because our model is trained solely for the task of synthesizing motion blur, its ``flow'' often looks irregular and inaccurate compared to optical flow algorithms, which are trained or designed to minimized end point error of with respect to scene motion. This difference manifests itself in a number of ways: our model assigns a near-zero ``flow'' to pixels in large flat regions of the image, because blurring a flat region looks identical to not blurring a flat region and so our training loss is agnostic in these flat regions. Also, our model attempts to model the motion of things like shadows, which optical flow algorithms are trained to ignore as they do not represent motion of the underlying physical object. This disconnect between apparent motion in an image and true motion in world geometry may explain why our optical flow baselines perform poorly on our task.
This difference between our model's learned ``flow'' and explicit optical flow techniques is analogous to prior work on learning monocular depth cues using defocus blur as a supervisory cue~\cite{Srinivasan2018}.

As our test-set performance demonstrates, our model performs well on diverse cases, including a variety of scene content, types of motion, duration of blurs, and amounts of blur in input frames. However, our model is limited in its inability to handle motions larger than those in the training dataset (32 pixels) and (similarly to other techniques) its inability to render nonlinear motion blur.

\newcommand{\flowwidth}{0.47\linewidth}
\newcommand{\flownumber}{20}
\begin{figure}  \centering
  \subfigure[Input images, averaged]{
  \label{fig:flow1}
  \includegraphics[width=\flowwidth]{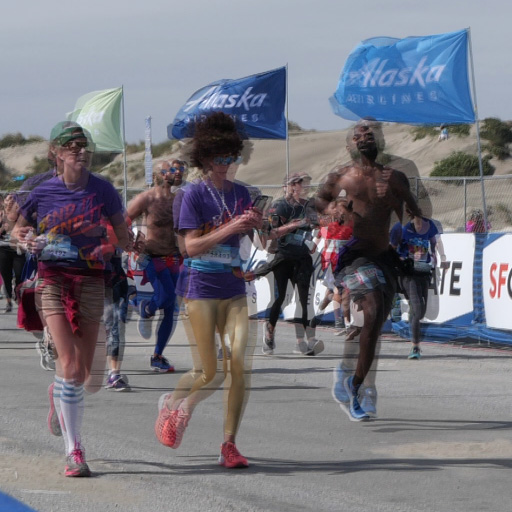}
  }
  \subfigure[\namemodel's $\Delta_1(x, y)$]{
  \label{fig:flow4}
  \includegraphics[width=\flowwidth]{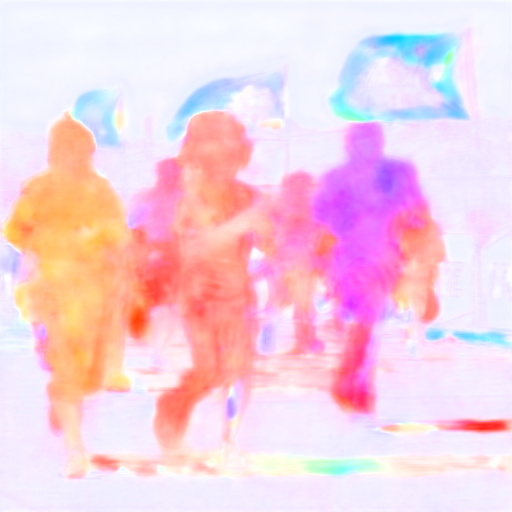}
  }
  \subfigure[\nameepic]{
  \label{fig:flow2}
  \includegraphics[width=\flowwidth]{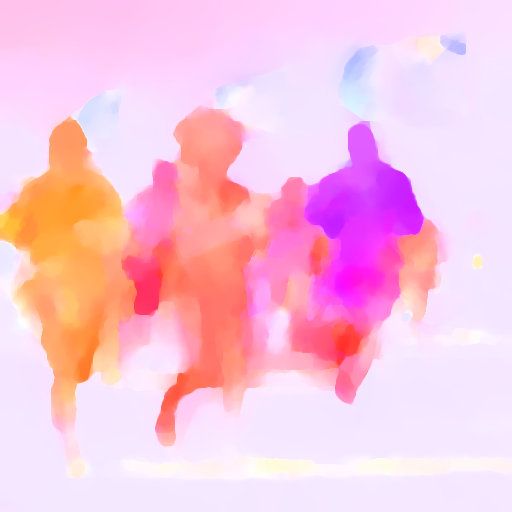}
  }
  \subfigure[\namepwc]{
  \label{fig:flow3}
  \includegraphics[width=\flowwidth]{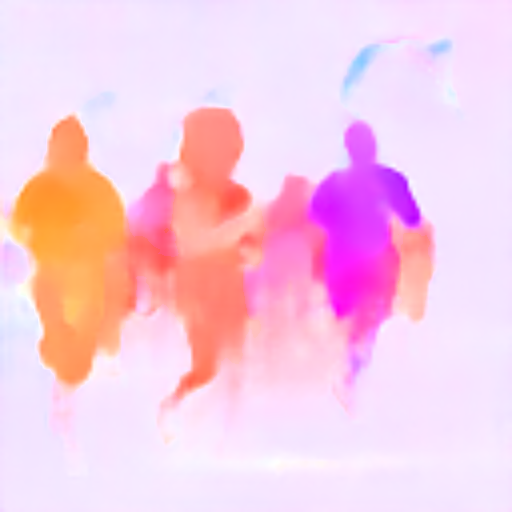}
  }
  \caption{
  A subset of our model's output can be visualized by using the endpoint of each pixel's predicted line as a flow vector. Here we render our model's ``flow field'' alongside two optical flow algorithms. Our ``flow fields'' tend to look irregular, highlighting the difference between training for accurate motion blur synthesis and training for accurate motion estimation.
  \label{fig:flowvis}
  }
\end{figure}

In the supplemental video we present results in which our system has been used to add motion blur to video sequences, by running on all pairs of adjacent video frames.

\section{Conclusion}
\label{sec:conclusion}

We have presented a technique for synthesizing motion blurred images from pairs of unblurred images.
As part of our neural network architecture we have proposed a novel line prediction layer, which is motivated by the optical properties of motion blur, and which is capable of producing accurate motion blur even when faced with occlusion and complex motion. We have described a strategy for using frame interpolation techniques to generate a large-scale synthetic dataset for use in training our motion blur synthesis model.
We additionally captured a ground truth test set of real motion blurred images with their corresponding input images, and with that we have demonstrated that our proposed model outperforms prior work in terms of accuracy and speed.
Our approach is fast, accurate, and uses readily available imagery from videos or ``bursts'' as input, and so provides a path for enabling motion blur manipulation in consumer photography applications, and for synthesizing the realistic training data needed by deblurring or motion estimation algorithms.

{\small
\bibliographystyle{ieee}
\bibliography{references}
}

\clearpage

\begin{appendices}

\section{Additional Results}

Because our synthetic dataset contains a validation set, we report performance of our model and its ablations in Table~\ref{table:synth_results}. We do not report the performance of our baseline techniques, as their performance on this synthetic data is unlikely to be meaningful when compared to our real test dataset, and also because some of our baselines needed to be run by the respective authors of each paper whom we did not wish to burden by requesting they process 15000 images in addition to our test set. In the table we see that the relative ordering of our model with respect to its ablations is consistent with their ordering in our test-set, though absolute performance is consistently higher.

\begin{table}[h]
\begin{center}
\begin{tabular}{l|ccc}
Algorithm              & PSNR & SSIM \\
\hline
\namedirect & 35.371 & 0.9854 \\
\namekpn    & 36.762 & 0.9873 \\
\nameavg    & 37.217 & 0.9866 \\
\namemodel  & 37.673 & 0.9881 
\end{tabular}
\caption{Performance of our model and its ablations on the validation set of our synthetic dataset.}
\label{table:synth_results}
\end{center}
\end{table}

See Figures~\ref{fig:results1}-\ref{fig:results5} for additional results on our real dataset, in which we compare our model against a set of ablations as well as a set of optical flow and video frame interpolation methods that could also be used to synthesize motion blurred images.

\renewcommand{\resultswidth}{0.32\linewidth}

\renewcommand{\resultname}{14_commute_day}
\begin{figure*}  \centering
  \subfigure[Input image 1]{
  \label{fig:results11}
  \includegraphics[width=\resultswidth]{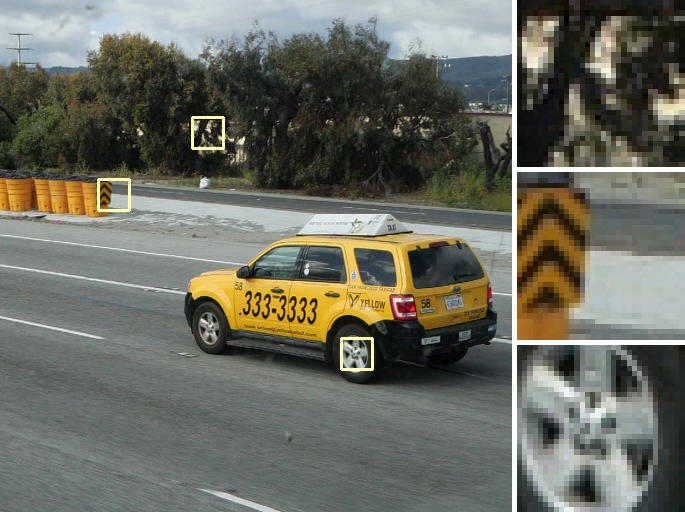}
  }
  \subfigure[Input image 2]{
  \label{fig:results12}
  \includegraphics[width=\resultswidth]{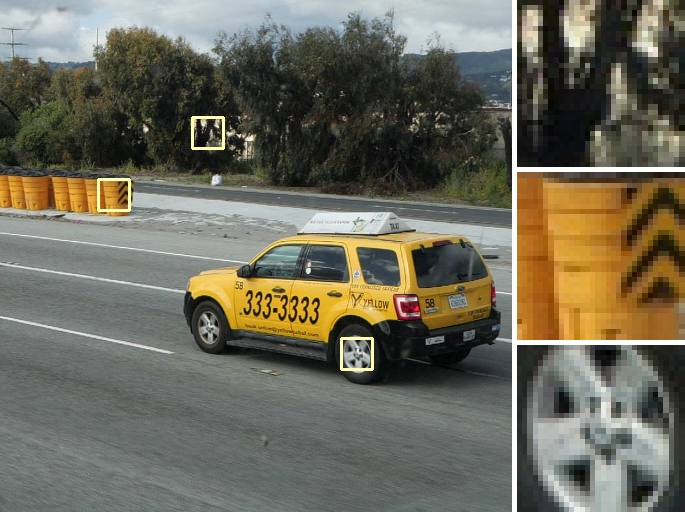}
  }
  \subfigure[Non-input intermediate frames]{
  \label{fig:results13}
  \includegraphics[width=\resultswidth]{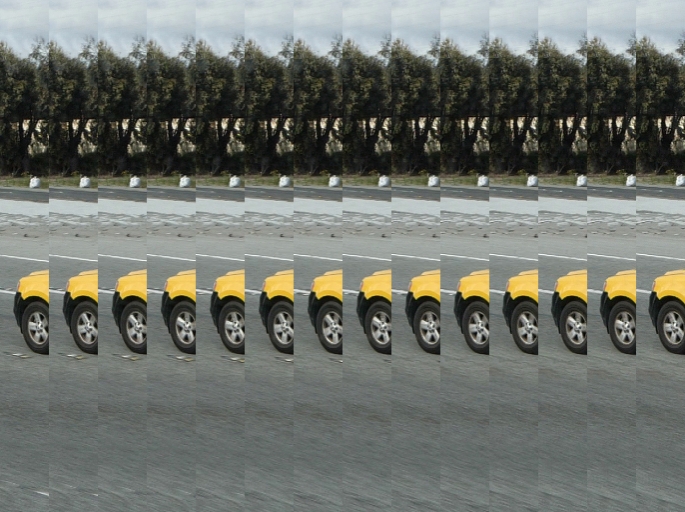}
  }
  \subfigure[Ground-truth motion blur]{
  \label{fig:results14}
  \includegraphics[width=\resultswidth]{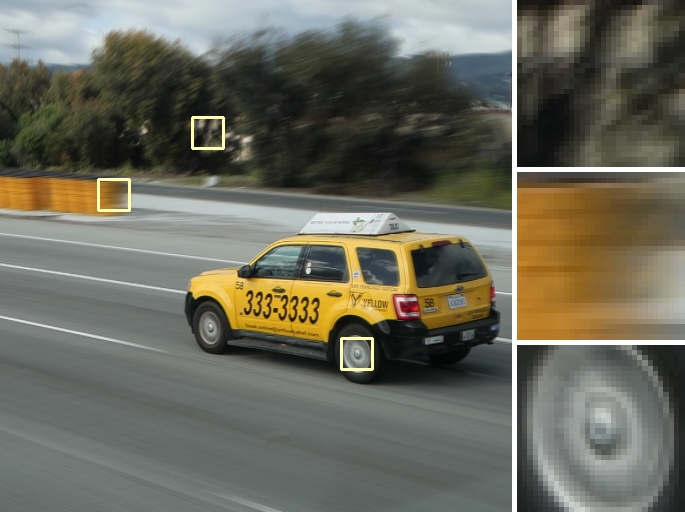}
  }
  \subfigure[\namepwc]{
  \includegraphics[width=\resultswidth]{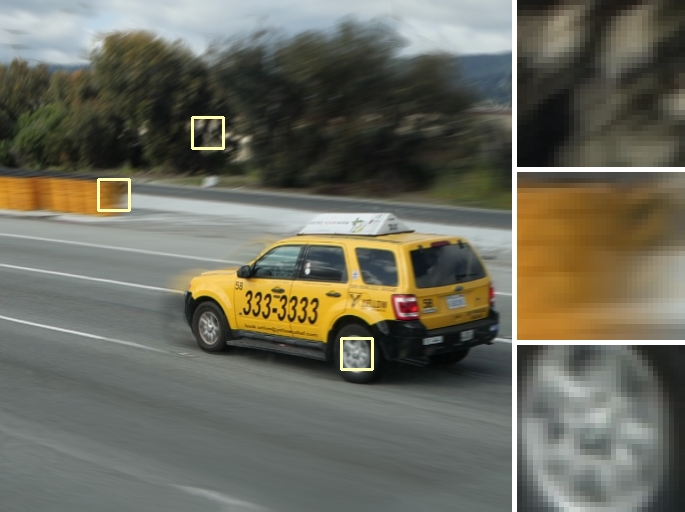}
  \label{fig:results15}
  }
  \subfigure[\nameepic]{
  \includegraphics[width=\resultswidth]{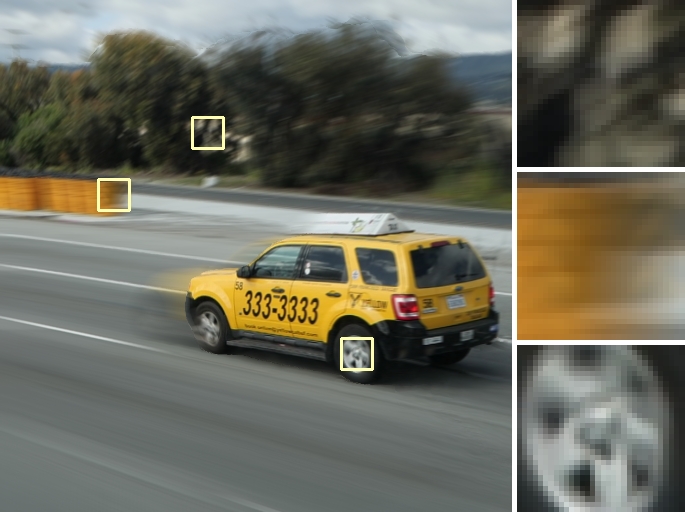}
  \label{fig:results16}
  }
  \subfigure[\namesepconv]{
  \includegraphics[width=\resultswidth]{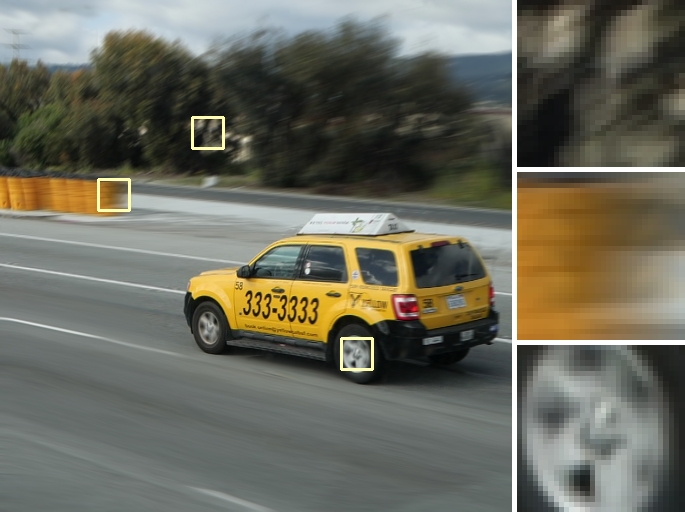}
  \label{fig:results17}
  }
  \subfigure[\nameslomo]{
  \includegraphics[width=\resultswidth]{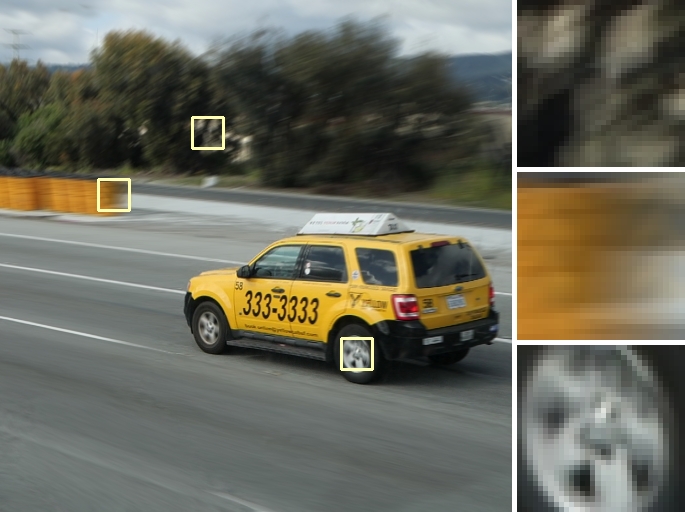}
  \label{fig:results18}
  }
  \subfigure[\namedirect]{
  \includegraphics[width=\resultswidth]{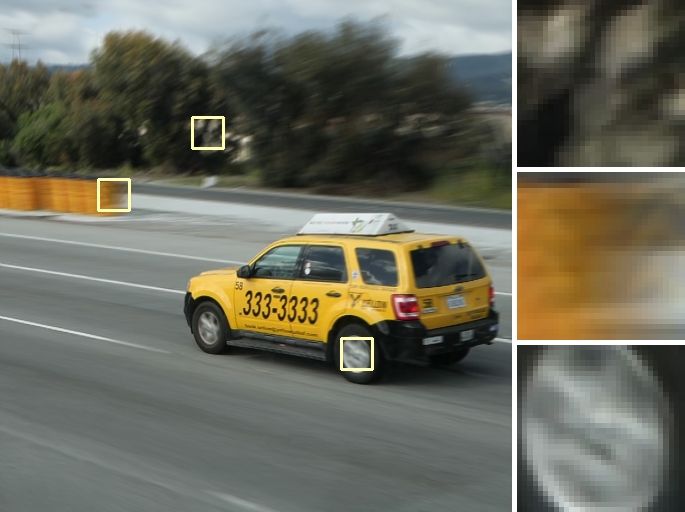}
  \label{fig:results19}
  }
  \subfigure[\nameavg]{
  \includegraphics[width=\resultswidth]{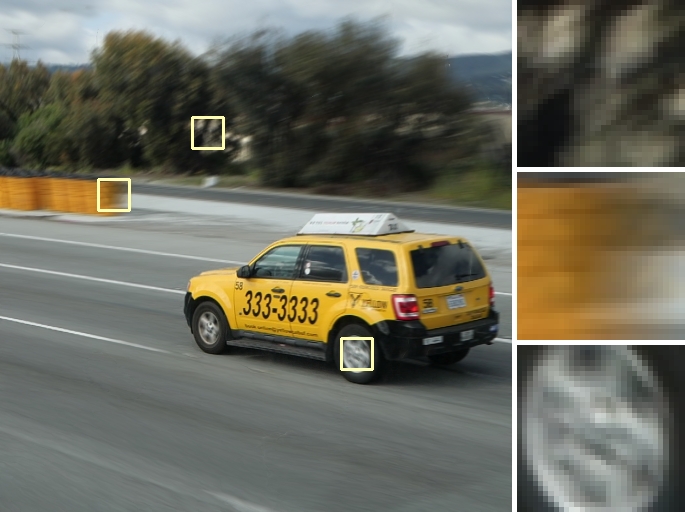}
  \label{fig:results19a}
  }
  \subfigure[\namekpn]{
  \includegraphics[width=\resultswidth]{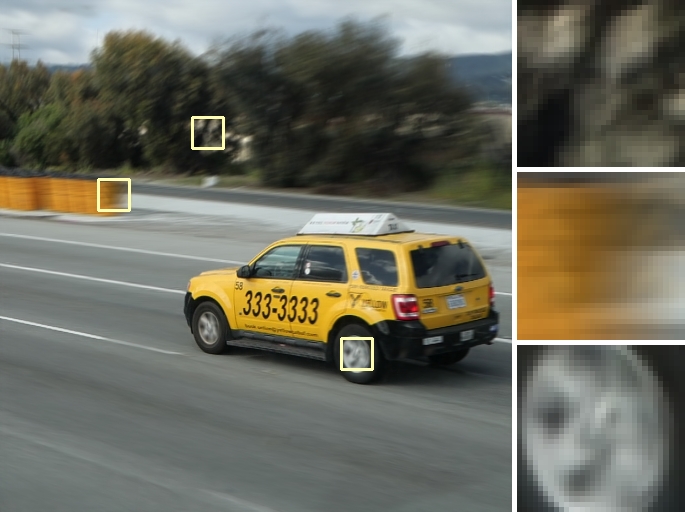}
  \label{fig:results19b}
  }
  \subfigure[\namemodel]{
  \includegraphics[width=\resultswidth]{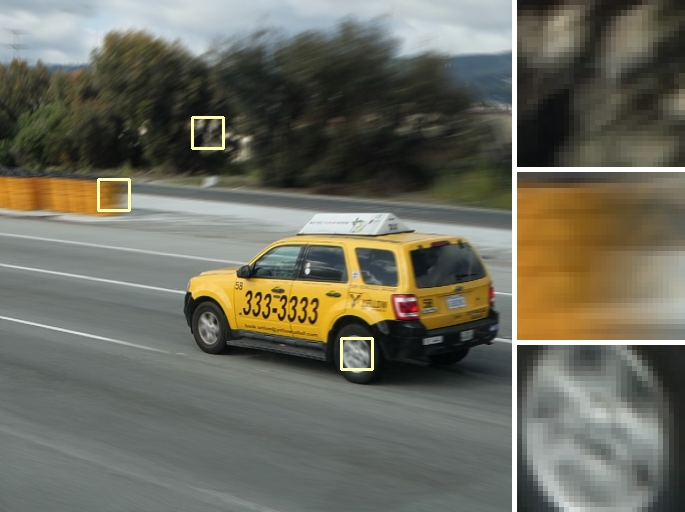}
  \label{fig:results19c}
  }
  \caption{
  Results for one scene from our test dataset.
  The ground truth image \subref{fig:results14} is the sum of the input images \subref{fig:results11} \& \subref{fig:results12} and of the frames between those two images \subref{fig:results13}.
  We programmatically select the three non-overlapping $32 \times 32$ sub-images with maximal variance across all frames in \subref{fig:results13} and present crops of those regions, rendered with nearest-neighbor interpolation and sorted by their $y$-coordinates. We compare our model \subref{fig:results19c} against four baselines \subref{fig:results15}-\subref{fig:results18}, and three ablations \subref{fig:results19}-\subref{fig:results19b}. Note that all techniques are unable to accurately blur the spinning wheel, which violates our model's and optical flow's assumption of linear motion.
  \label{fig:results1}
  }
\end{figure*}

\renewcommand{\resultname}{01_trees}
\begin{figure*}[p]  \centering
  \subfigure[Input image 1]{
  \includegraphics[width=\resultswidth]{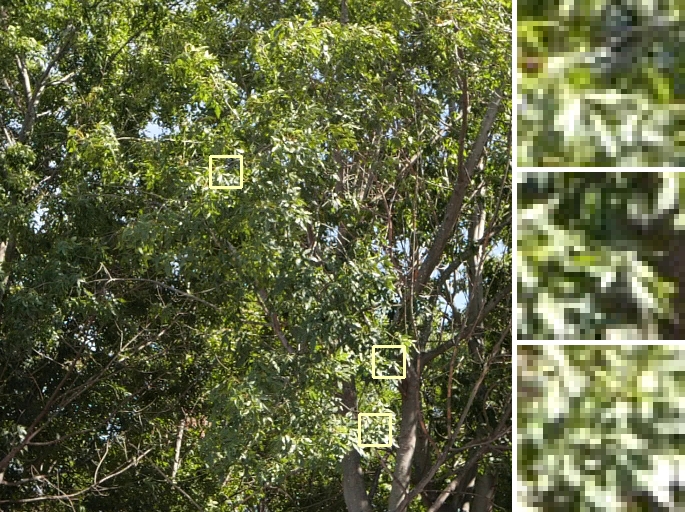}
  }
  \subfigure[Input image 2]{
  \includegraphics[width=\resultswidth]{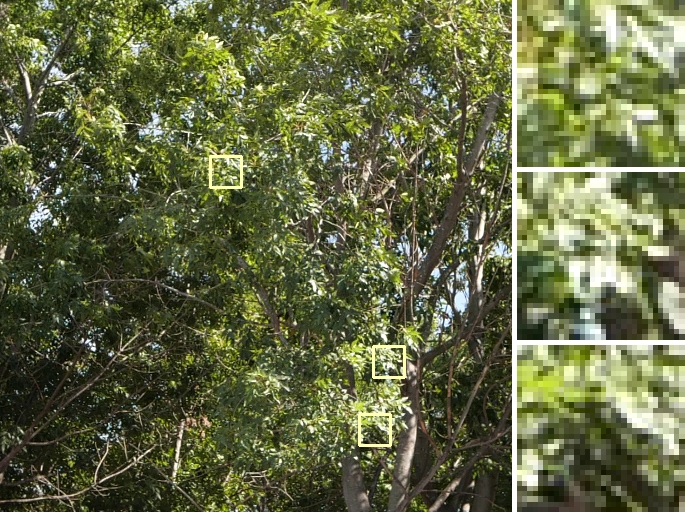}
  }
  \subfigure[Non-input intermediate frames]{
  \includegraphics[width=\resultswidth]{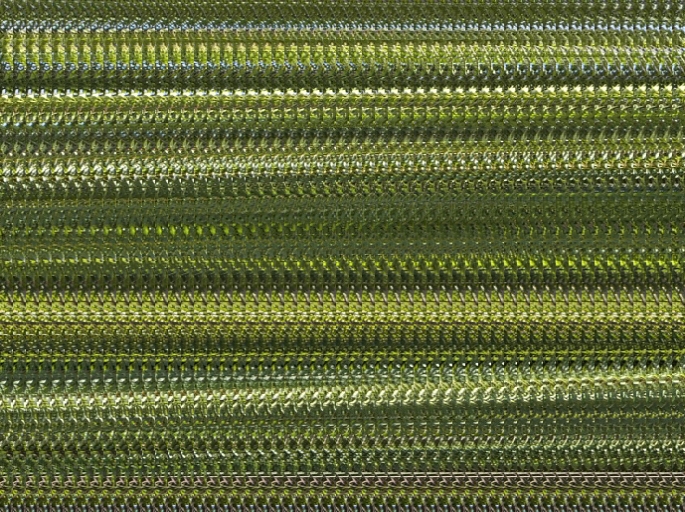}
  }
  \subfigure[Ground-truth motion blur]{
  \includegraphics[width=\resultswidth]{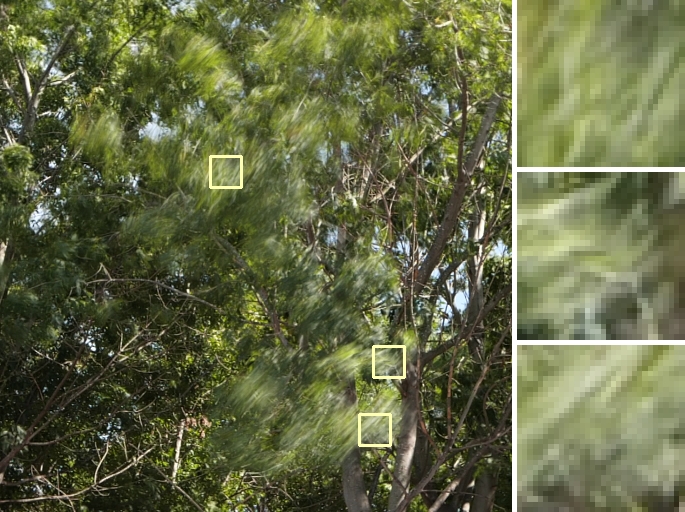}
  }
  \subfigure[\namepwc]{
  \includegraphics[width=\resultswidth]{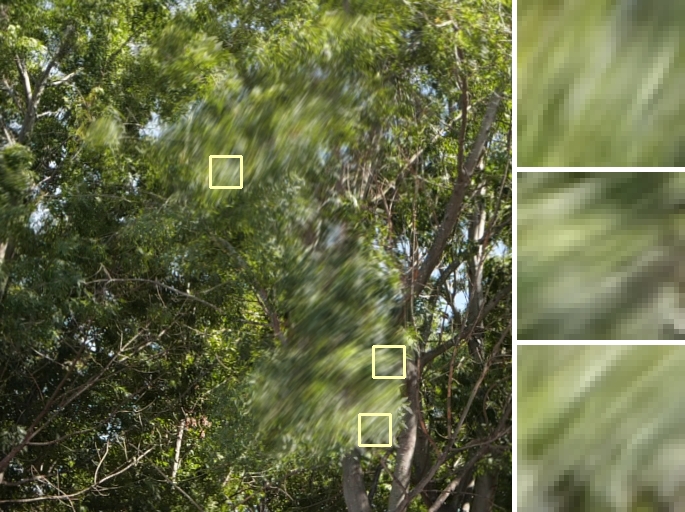}
  }
  \subfigure[\nameepic]{
  \includegraphics[width=\resultswidth]{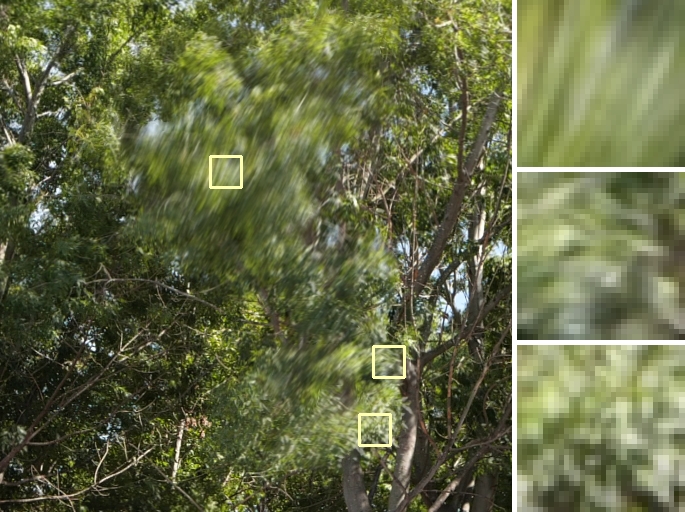}
  }
  \subfigure[\namesepconv]{
  \includegraphics[width=\resultswidth]{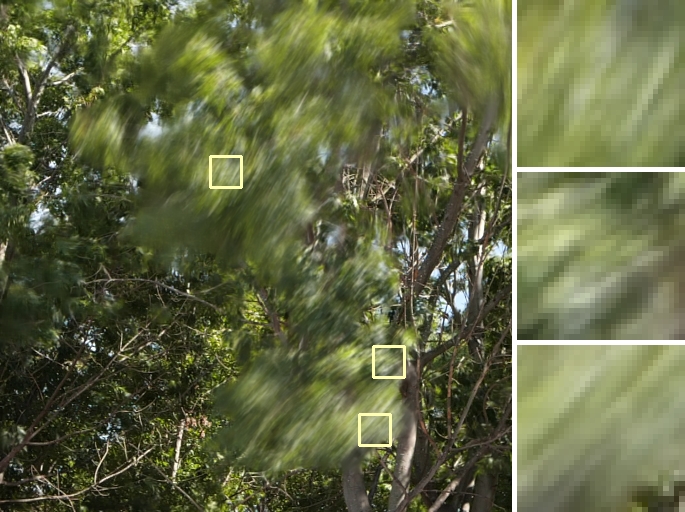}
  }
  \subfigure[\nameslomo]{
  \includegraphics[width=\resultswidth]{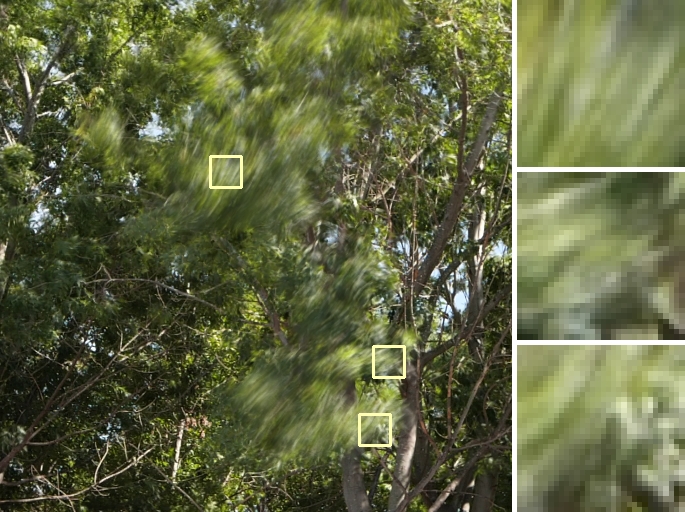}
  }
  \subfigure[\namedirect]{
  \includegraphics[width=\resultswidth]{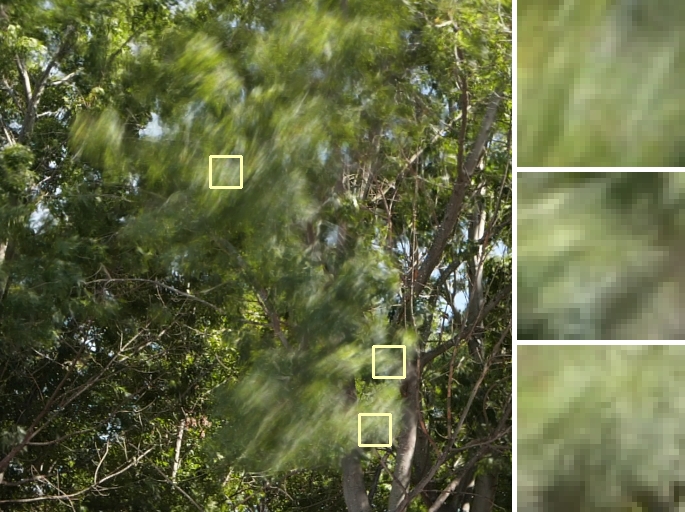}
  }
  \subfigure[\nameavg]{
  \includegraphics[width=\resultswidth]{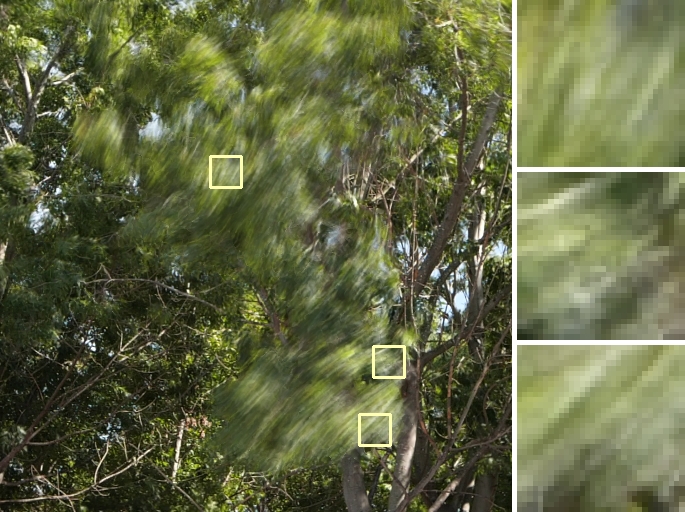}
  }
  \subfigure[\namekpn]{
  \includegraphics[width=\resultswidth]{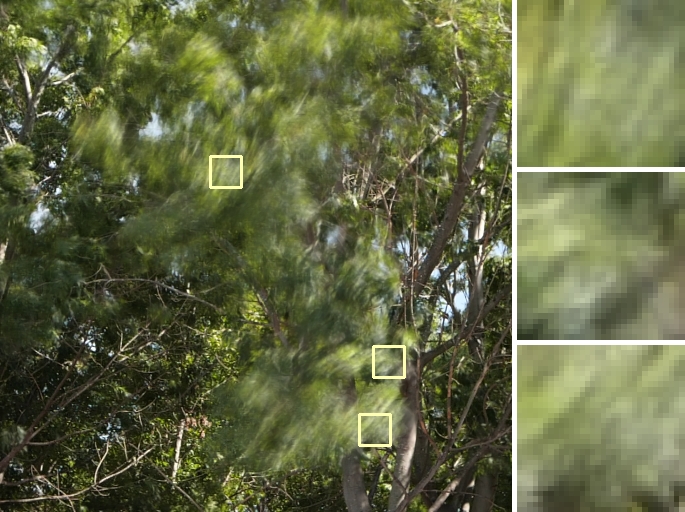}
  }
  \subfigure[\namemodel]{
  \includegraphics[width=\resultswidth]{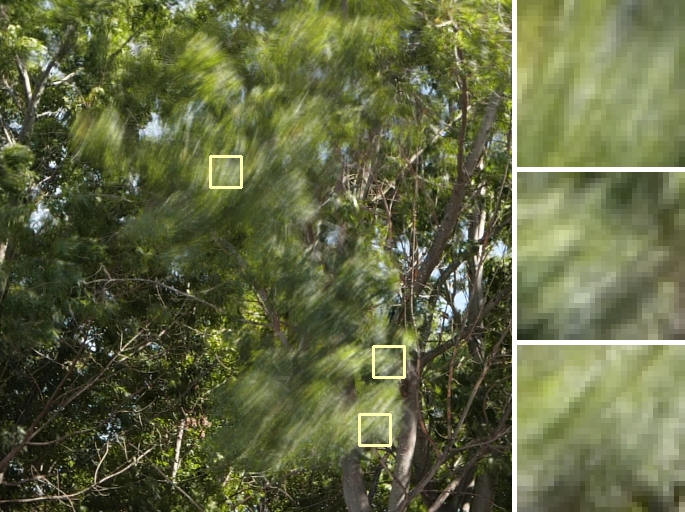}
  }
  \caption{Additional results in the same format as Figure~\ref{fig:results1}.
  \label{fig:results3}
  }
\end{figure*}

\renewcommand{\resultname}{10_intersection}
\begin{figure*}[p]  \centering
  \subfigure[Input image 1]{
  \includegraphics[width=\resultswidth]{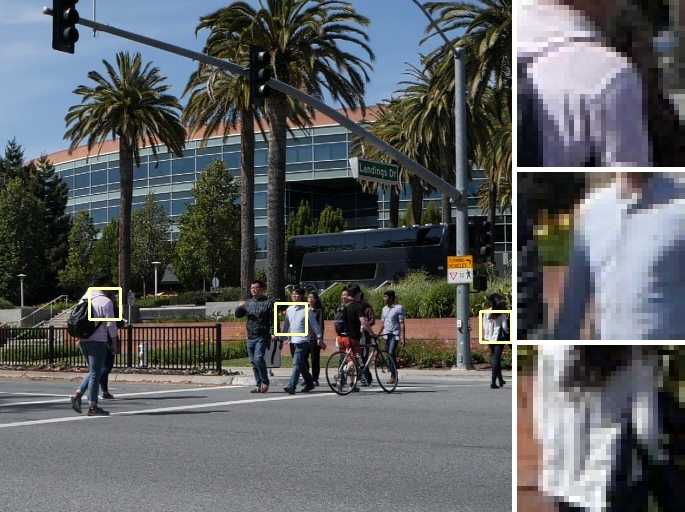}
  }
  \subfigure[Input image 2]{
  \includegraphics[width=\resultswidth]{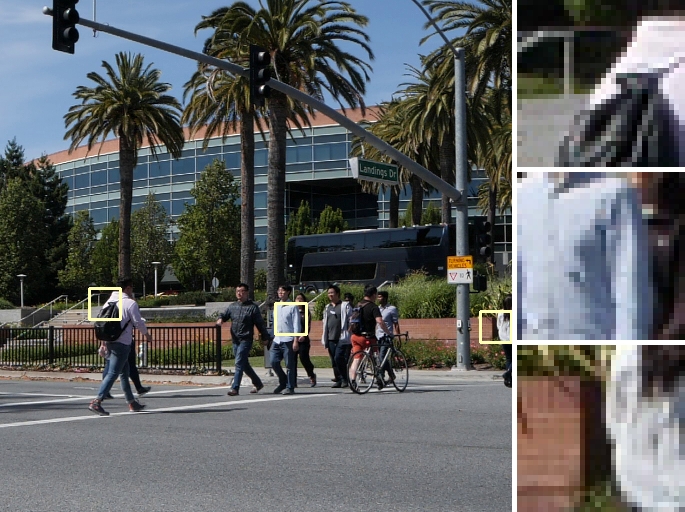}
  }
  \subfigure[Non-input intermediate frames]{
  \includegraphics[width=\resultswidth]{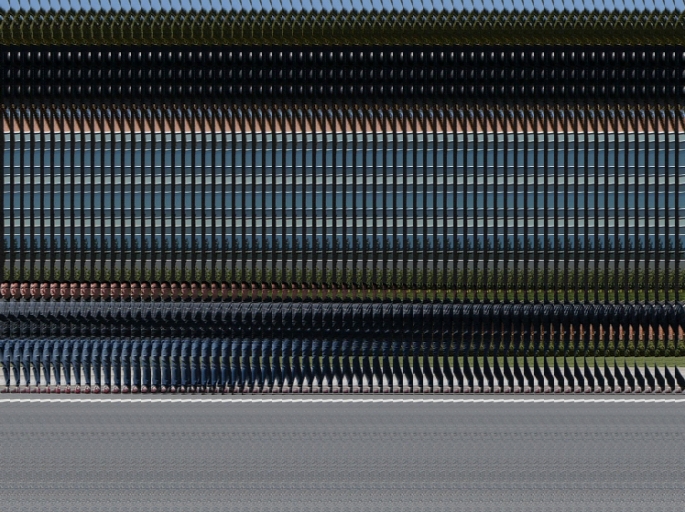}
  }
  \subfigure[Ground-truth motion blur]{
  \includegraphics[width=\resultswidth]{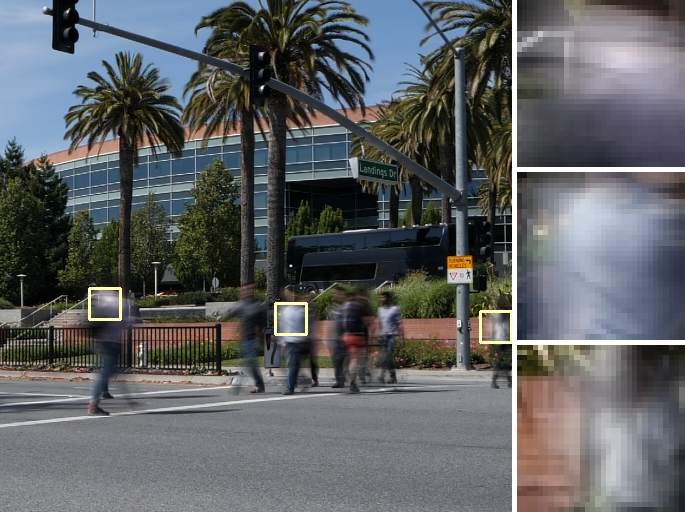}
  }
  \subfigure[\namepwc]{
  \includegraphics[width=\resultswidth]{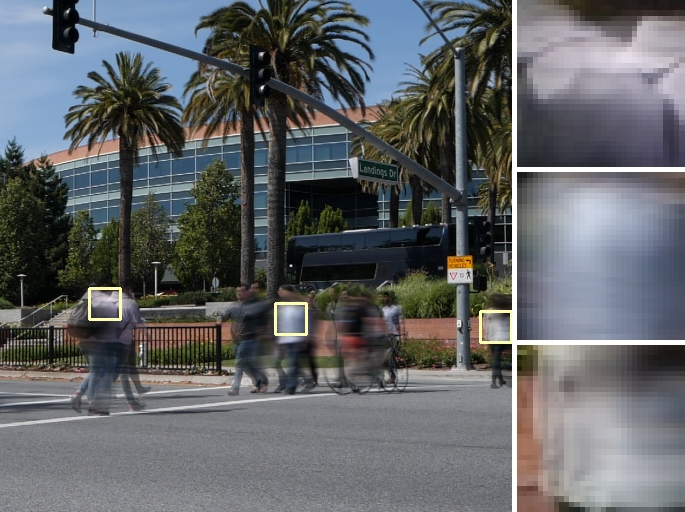}
  }
  \subfigure[\nameepic]{
  \includegraphics[width=\resultswidth]{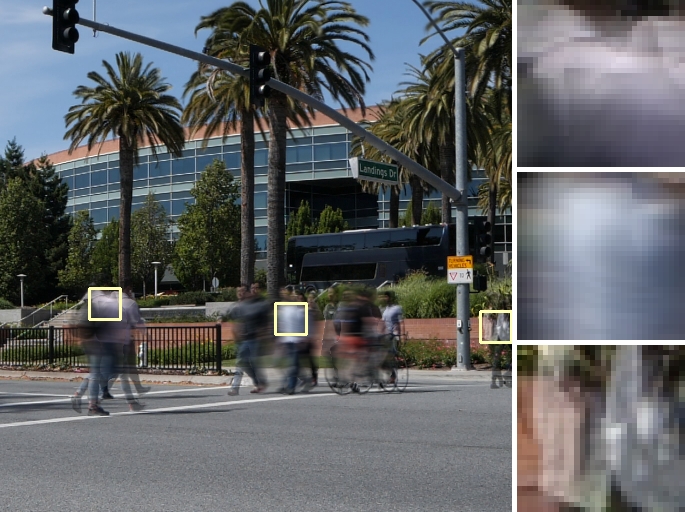}
  }
  \subfigure[\namesepconv]{
  \includegraphics[width=\resultswidth]{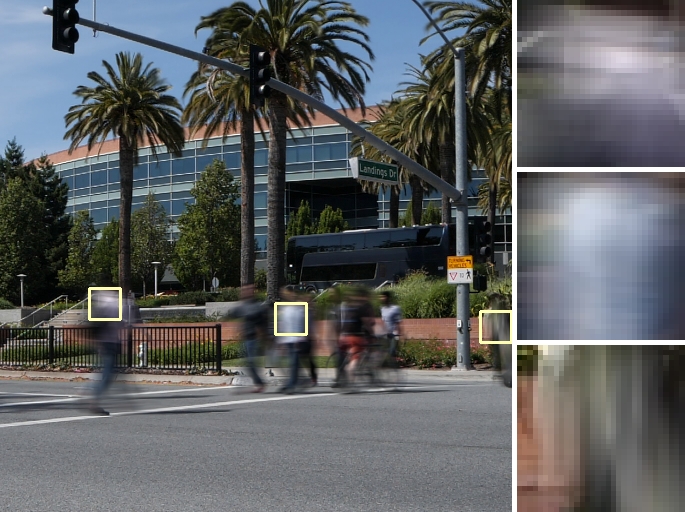}
  }
  \subfigure[\nameslomo]{
  \includegraphics[width=\resultswidth]{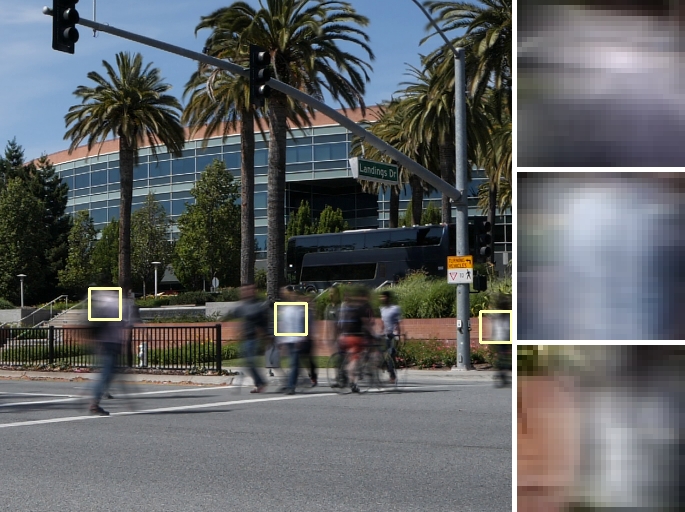}
  }
  \subfigure[\namedirect]{
  \includegraphics[width=\resultswidth]{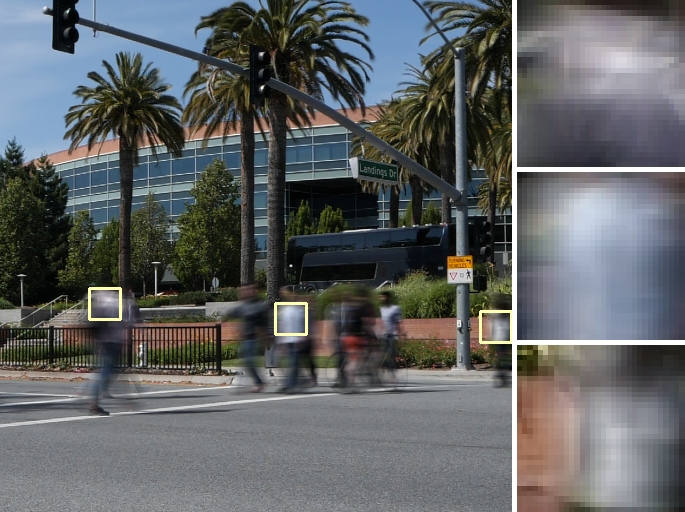}
  }
  \subfigure[\nameavg]{
  \includegraphics[width=\resultswidth]{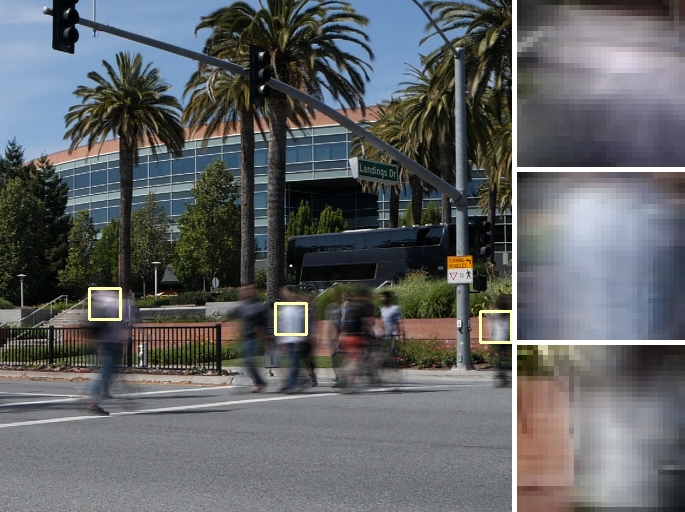}
  }
  \subfigure[\namekpn]{
  \includegraphics[width=\resultswidth]{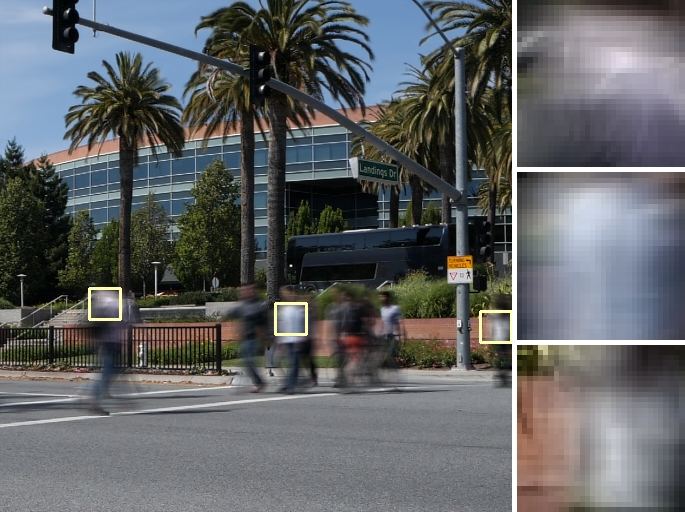}
  }
  \subfigure[\namemodel]{
  \includegraphics[width=\resultswidth]{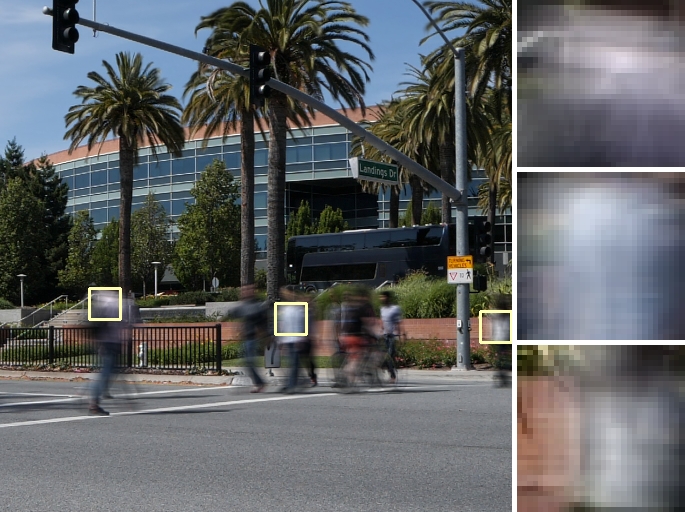}
  }
  \caption{Additional results in the same format as Figure~\ref{fig:results1}.
  \label{fig:results4}
  }
\end{figure*}

\renewcommand{\resultname}{20_bay_breakers}
\begin{figure*}[p]  \centering
  \subfigure[Input image 1]{
  \includegraphics[width=\resultswidth]{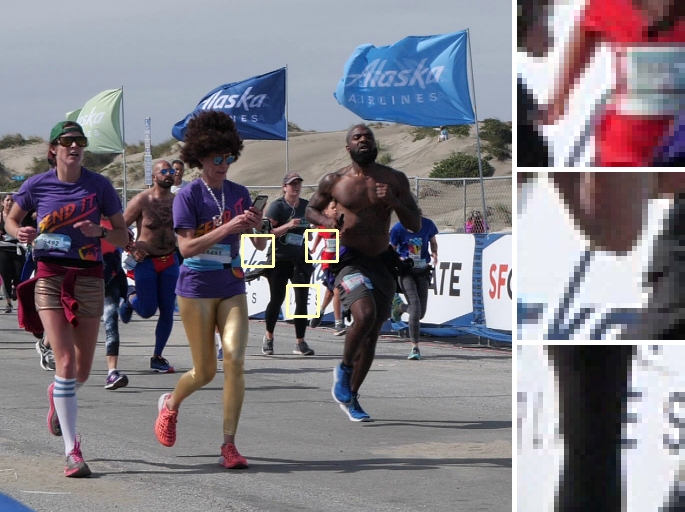}
  }
  \subfigure[Input image 2]{
  \includegraphics[width=\resultswidth]{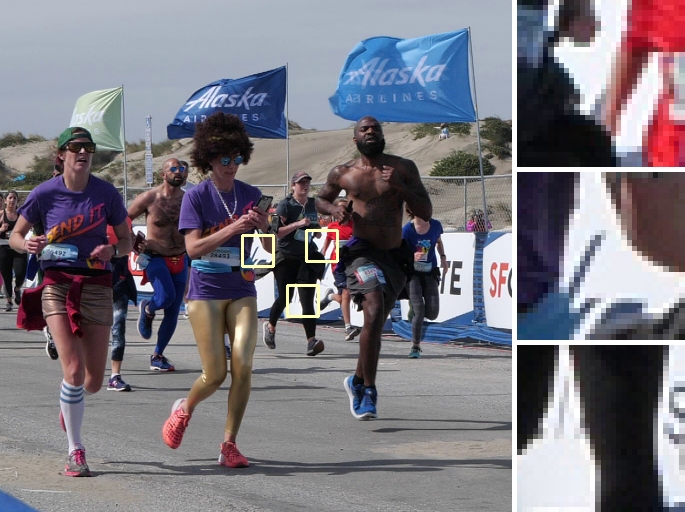}
  }
  \subfigure[Non-input intermediate frames]{
  \includegraphics[width=\resultswidth]{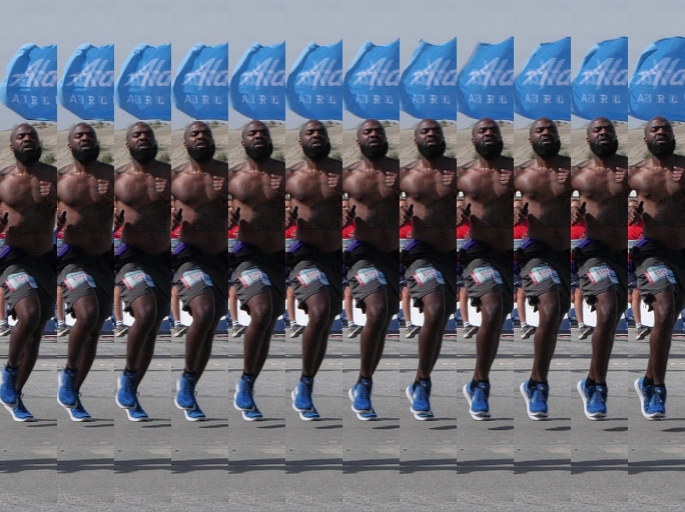}
  }
  \subfigure[Ground-truth motion blur]{
  \includegraphics[width=\resultswidth]{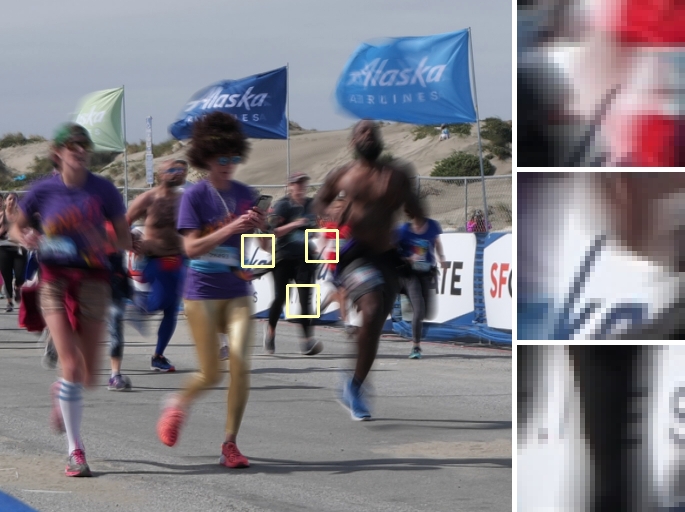}
  }
  \subfigure[\namepwc]{
  \includegraphics[width=\resultswidth]{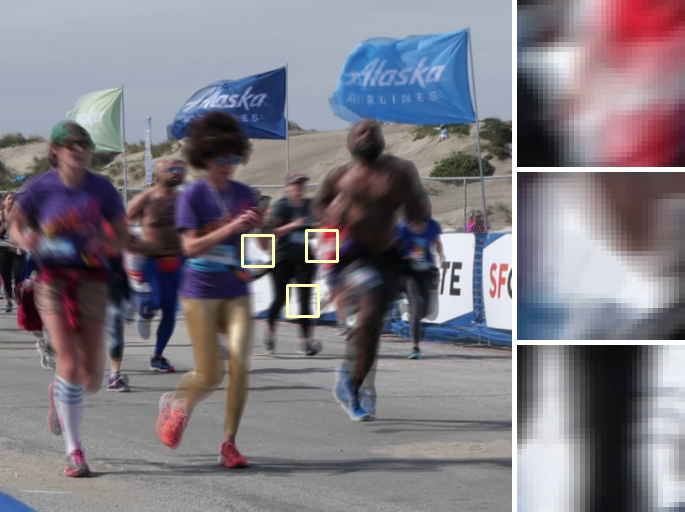}
  }
  \subfigure[\nameepic]{
  \includegraphics[width=\resultswidth]{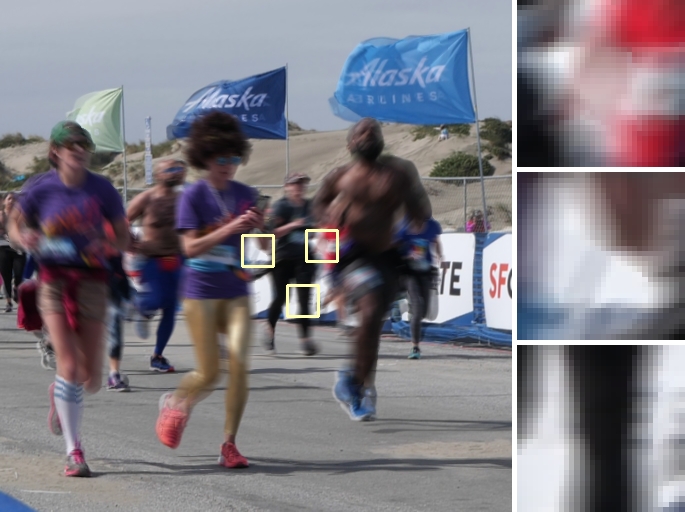}
  }
  \subfigure[\namesepconv]{
  \includegraphics[width=\resultswidth]{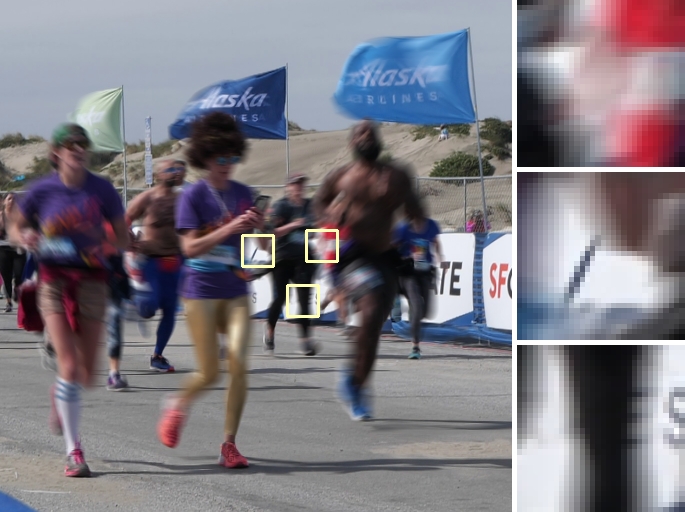}
  }
  \subfigure[\nameslomo]{
  \includegraphics[width=\resultswidth]{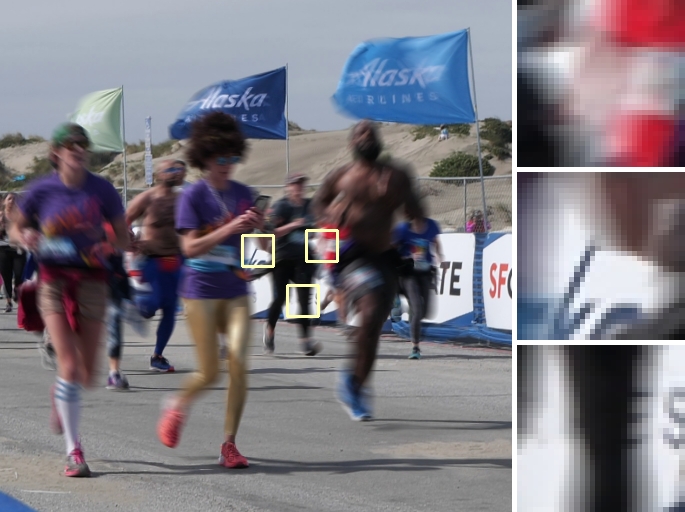}
  }
  \subfigure[\namedirect]{
  \includegraphics[width=\resultswidth]{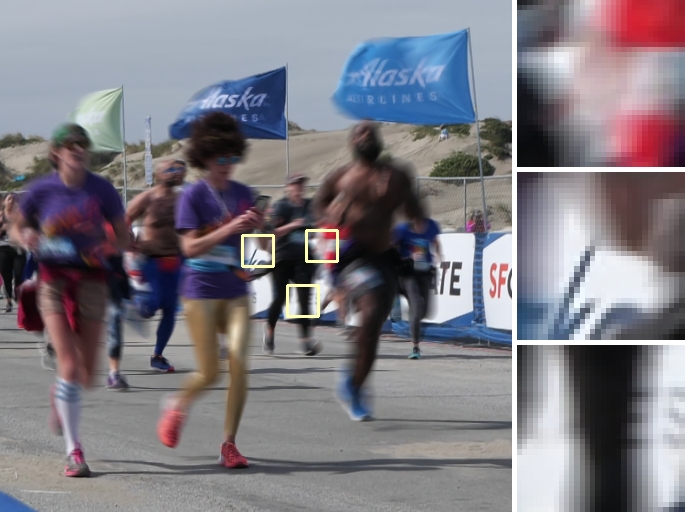}
  }
  \subfigure[\nameavg]{
  \includegraphics[width=\resultswidth]{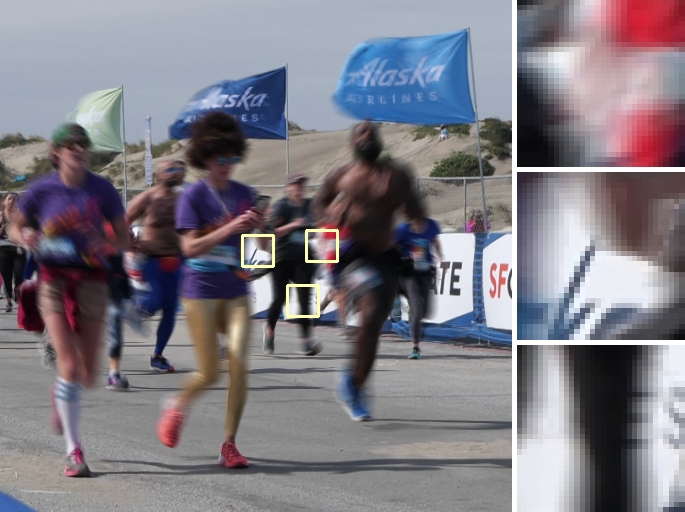}
  }
  \subfigure[\namekpn]{
  \includegraphics[width=\resultswidth]{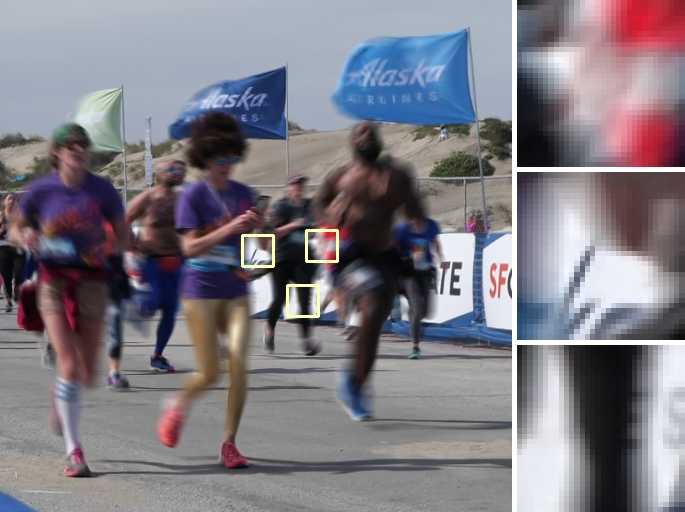}
  }
  \subfigure[\namemodel]{
  \includegraphics[width=\resultswidth]{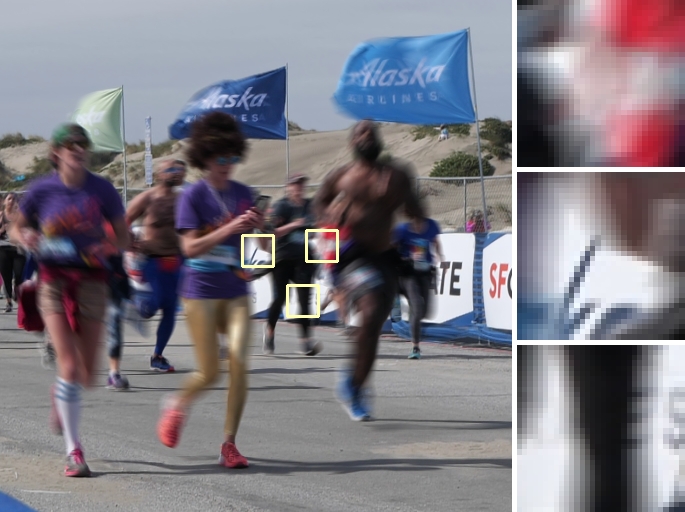}
  }
  \caption{Additional results in the same format as Figure~\ref{fig:results1}.
  \label{fig:results5}
  }
\end{figure*}

\end{appendices}

\end{document}